\documentclass[10pt,twocolumn,letterpaper]{article}

\usepackage{iccv}
\usepackage{times}
\usepackage{epsfig}
\usepackage{graphicx}
\usepackage{amsmath}
\usepackage{amssymb}

\usepackage[dvipsnames]{xcolor}
\usepackage{subcaption}
\usepackage{caption}
\usepackage[labelfont=bf]{caption}
\usepackage{float}
\usepackage{booktabs}
\usepackage{colortbl}
\usepackage{adjustbox}
\usepackage{microtype}
\usepackage{multirow}

\usepackage[colorlinks=true, citecolor=citecolor, linkcolor=linkcolor, breaklinks=true,bookmarks=false]{hyperref}
\definecolor{citecolor}{HTML}{0071BC}
\definecolor{linkcolor}{HTML}{ED1C24}
\usepackage{verbatim}
\usepackage[capitalize]{cleveref}
\usepackage{authblk} 

\iccvfinalcopy 

\def\iccvPaperID{****} 

\ificcvfinal\fi

\title{FeatEnHancer: Enhancing Hierarchical Features for Object Detection and Beyond Under Low-Light Vision}

\author[1,2]{Khurram Azeem Hashmi}
\author[2]{Goutham Kallempudi}
\author[1,2]{Didier Stricker}
\author[1,2]{Muhammamd Zeshan Afzal}
\affil[1]{DFKI - German Research Center for Artificial Intelligence}
\affil[2]{MindGarage, RPTU Kaiserslautern}
\date{}                     
\begin{document}

\twocolumn[{%
\renewcommand\twocolumn[1][]{#1}%
\maketitle
\begin{center}
    \centering
    \captionsetup{type=figure}
    \includegraphics[width=\textwidth]{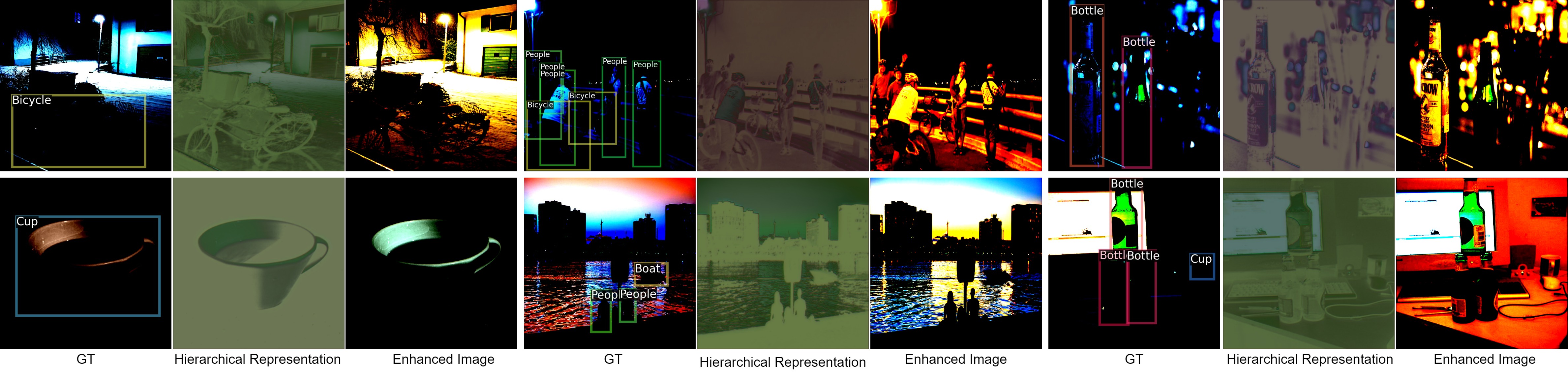}
    \captionof{figure}{\textbf{Learned hierarchical representation and enhanced image from our FeatEnHancer.} We train our FeatEnHancer on a downstream object detection task and visualize these images from the validation set. These maps and enhanced images show that despite producing less visually appealing images, our model enhances task-related features. Best viewed on the screen.}
    \label{fig:intro_figure}
\end{center}%
}]

\ificcvfinal\thispagestyle{empty}\fi

\begin{abstract}
    Extracting useful visual cues for the downstream tasks is especially challenging under low-light vision. Prior works create enhanced representations by either correlating visual quality with machine perception or designing illumination-degrading transformation methods that require pre-training on synthetic datasets. We argue that optimizing enhanced image representation pertaining to the loss of the downstream task can result in more expressive representations. Therefore, in this work, we propose a novel module, FeatEnHancer, that hierarchically combines multiscale features using multi-headed attention guided by task-related loss function to create suitable representations. Furthermore, our intra-scale enhancement improves the quality of features extracted at each scale or level, as well as combines features from different scales in a way that reflects their relative importance for the task at hand. FeatEnHancer is a general-purpose plug-and-play module and can be incorporated into any low-light vision pipeline. We show with extensive experimentation that the enhanced representation produced with FeatEnHancer significantly and consistently improves results in several low-light vision tasks, including dark object detection (+5.7 mAP on ExDark), face detection (+1.5 mAP on DARK FACE), nighttime semantic segmentation (+5.1 mIoU on ACDC), and video object detection (+1.8 mAP on DarkVision), highlighting the effectiveness of enhancing hierarchical features under low-light vision.
\end{abstract}

\section{Introduction}
\label{sec:intorduction}

Recent remarkable advancements in high-level vision tasks have shown that given a high-quality image, current vision backbone networks~\cite{ResNet, Res2Net_PAMI_2021, Vit, Swin_ICCV_21, Swin_V2_CVPR22}, object detectors~\cite{yolov3, retinaNet, FasterRCNN, Mask_RCNN_ICCV_17, CascadeRCNN,cascade_Rcnn_pami, SparseRCNN, DETR, deformable_detr, dino_detr_ICLR_23, featurized-query-rcnn} and semantic segmentation models~\cite{sem_seg_FCN_CVPR_15, segmenter_iccv21,UpperNEt_ECCV_18, deeplabv3_ECCV_2018, segformer_neurips_21}
can effectively learn desired features to perform vision tasks.~Similarly, modern low-light image enhancement (LLIE) methods~\cite{raus, kind, mebbln, enlightengan,zero-dce, zero-dce++} are capable of transforming a low-light image into a visual-friendly representation. However, a naive combination of the two brings sub-optimal gains when it comes to  high-level vision tasks under low-light vision.

This work explores the underlying reasons for the low performance of the combination of LLIE with high-level vision methods and observes the following limitations:~1) Although existing LLIE methods push the envelope of visual perception for human eyes, they do not align with vision backbone networks~\cite{ResNet, Vit, Res2Net_PAMI_2021, Swin_ICCV_21, Swin_V2_CVPR22} due to lack of multi-scale features.~For instance, it is likely that the enhancement method increases brightness in some regions.~However, it simultaneously corrupts the edges and texture information of objects. 2) The pixel distribution among different low-light images may have huge variance owing to the disparity in less illuminated environments~\cite{zero-dce, zero-dce++,STAR_ICCV}.~This increases intra-class variance in some cases (see Fig.~\ref{fig:qual_exdark}, where only one bicycle is recognized by~\cite{zero-dce} instead of two bicycles in the ground-truth).~3) Current LLIE approaches~\cite{mebbln, zero-dce, zero-dce++, enlightengan, raus, URetinex-Net, kind} employ enhancement loss functions to optimize the enhancement networks. These loss functions compel the network to attend to all pixels equally, lacking the learning of informative details necessary for high-level downstream vision tasks such as object pose and shape for object detection.~Furthermore, to train these enhancement networks, most of them~\cite {raus, mebbln, kind, URetinex-Net} require a set of high-quality images, which is hardly available in a real-world setting.

Motivated by these observations and inspired by recent developments in LLIE~\cite{zero-dce, zero-dce++} and vision-based backbone networks~\cite{Res2Net_PAMI_2021, Swin_ICCV_21, Swin_V2_CVPR22}, this paper aims to bridge the gap by exploring an end-to-end trainable recipe that jointly optimizes the enhancement and downstream task objectives in a single network.~To this end, we present FeatEnHancer, a general-purpose feature enhancer that learns to enrich multi-scale hierarchical features favourable for downstream vision tasks in a low-light setting. An example of learned hierarchical representation and the enhanced image is illustrated in Fig.~\ref{fig:intro_figure}. 

In particular, our FeatEnHancer first downsamples a low-light RGB input image to construct multi-scale hierarchical representations. Subsequently, these representations are fed to our Feature Enhancement Network (FEN), which is a deep convolutional network, employed to enrich intra-scale semantic representations. Note that the parameters of FEN can be adjusted through task-related loss functions, which pushes the FEN to only enhance the task-related features. This multi-scale learning allows the network to enhance both global and local information from higher and lower-resolution features, respectively.~Once the enhanced representations on different scales are obtained, the remaining obstacle is to fuse them effectively.~To achieve, this, we select two different strategies to capture both global and local information from higher and lower-resolution features. First, to merge high-resolution features, inspired by multi-head attention in~\cite{attn_all_need_Neurips_17}, we design a Scale-aware Attentional Feature Aggregation (SAFA) method that jointly attends information from different scales. Second, for lower-resolution features, the skip connection~\cite{ResNet} scheme is adopted to merge the enhanced representation from SAFA to lower-resolution features. With these jointly learned hierarchical features, our FeatEnHancer provides semantically powerful representations which can be exploited by advanced methods such as feature pyramid networks~\cite{FPN} for object detection~\cite{FasterRCNN} and instance segmentation~\cite{Mask_RCNN_ICCV_17}, or UNet~\cite{unet} for semantic segmentation~\cite{sem_seg_FCN_CVPR_15}.


The main contributions of this work can be summarized as follows:
\begin{enumerate}
    \item We propose FeatEnHancer, a novel module that enhances hierarchical features to boost downstream vision tasks under low-light vision.~Our intra-scale feature enhancement and scale-aware attentional feature aggregation schemes are aligned with vision backbone networks and produce powerful semantic representations. FeatEnHancer is a general-purpose plug-and-play module that can be trained end-to-end with any high-level vision task.\vspace{-5pt}
    \item To the best of our knowledge, this is the first work that fully exploits multi-scale hierarchical features in low-light scenarios and generalizes to several downstream vision tasks such as object detection, semantic segmentation, and video object detection. \vspace{-5pt}
    \item Extensive experiments on four different downstream visions tasks covering both images and videos demonstrate that our method brings consistent and significant improvements over baselines, LLIE methods and task-specific state-of-the-art approaches. \vspace{-5pt}
\end{enumerate}


\section{Related Work}
\label{sec:related_work}
\subsection{Enhancing Low-Light Images}
\label{sec:LLIE}
Deep learning-based LLIE methods focus on improving the visual quality of low-light images that satisfies human visual perception~\cite{land1986alternative, jobson1997multiscale}. Most~LLIE approaches~\cite{mebbln, raus, URetinex-Net, kind} operate under a supervised learning paradigm, requiring paired data during training.~Unsupervised GAN-based methods~\cite{enlightengan} eliminate the need for paired data during the training.~However,~their performance relies on the careful choice of unpaired data.~Recently, zero-reference methods~\cite{zero-dce, zero-dce++, STAR_ICCV} discard the need for both paired and unpaired data to enhance low-light images by designing a set of non-reference loss functions. Inspired by these recent developments, this
work aims to bridge low-light enhancement and downstream vision tasks (such as object detection~\cite{maet, exdark-dataset, DARK_FACE_dataset_1}, semantic segmentation~\cite{see_understand_dark_22_ACMM, ACDC_dataset_ICCV_21}, and video object detection~\cite{DarkVision_dataset_arxiv_23}) by enhancing multi-scale hierarchical features without needing paired or unpaired data to boost performance.

\vspace{-5 pt}
\subsection{Enhancing Low-Light for Downstream Vision Tasks}
\label{sec:lliv}
These approaches consider machine perception as the criteria for success while enhancing images to improve downstream vision tasks.~One obvious way to achieve this goal is to apply the LLIE methods as an initial step~\cite{ForkGAN_ECCV_2020, zero-dce}. However, this leads to unsatisfactory results (see Table~\ref{tab:results_exdark},~\ref{tab:results_acdc}, and~\ref{tab:results_dakrvision}). Recently, another line of work has explored end-to-end pipelines, optimizing both enhancement and individual tasks during training, and our work follows the same spirit.

\noindent\textbf{Face detection.}~Liang \etal~\cite{REG_face_22} propose an effective information extraction scheme from low-light images by exploiting multi-exposure generation.~Furthermore, bi-directional domain adaptation~\cite{HLA-Face_CVPR_21, HLA-Face_PAMI_23} and parallel architecture that jointly performs enhancement and detection~\cite{pia} are presented to advance the research.~However, these approaches are carefully designed to tackle face detection~\cite{DARK_FACE_dataset_1, DARKFACE_dataset_2} only and deliver minor improvements when applied to generic object detection~\cite{HLA-Face_PAMI_23}.~Contrarily, our FeatEnHancer is a general-purpose module. It significantly improves several downstream vision tasks.~Hence, we refrain from comparing our method to architectures only evaluated for face detection.\\

\begin{figure*}
\begin{center}
\includegraphics[width=\linewidth]{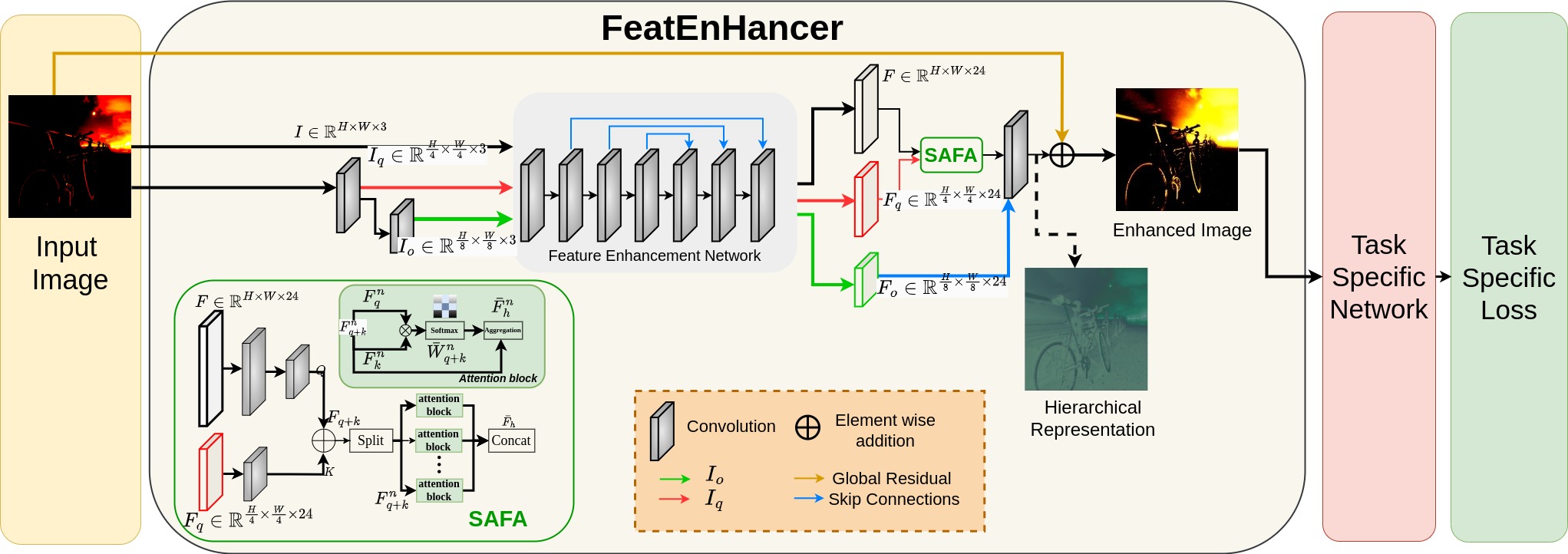}
\end{center}
   \caption{\textbf{Network architecture of the proposed FeatEnHancer employed in a downstream vision task}. Our FeatEnHancer takes a low-light image and adaptively boosts its semantic representation by enriching task-related hierarchical features. Zoom in for the best view.}
\label{fig:featenhancer}
\vspace{-10pt}
\end{figure*}

\vspace{-13pt}
\noindent\textbf{Dark object detection.} 
Dark (low-light) object detection~\cite{maet,image_adaptive_YOLO_AAAI_22} methods have emerged recently, thanks to the real-world low illumination datasets~\cite{exdark-dataset, nod}.~IA-YOLO~\cite{image_adaptive_YOLO_AAAI_22} introduces a convolutional neural network~(CNN)-based parameter predictor that learns the optimal configuration for the filters employed in the differential image processing module.~Most related to our work is MAET~\cite{maet}, which investigates the physical noise model and image signal processing (ISP) pipeline under low illumination and learns the model to predict degradation parameters and object features.~To avoid feature entanglement,~they impose orthogonal tangent regularity to penalize cosine similarity between objects and degrading features.~However, owing to the weather-specific hyperparameters in~\cite{image_adaptive_YOLO_AAAI_22} and degradation parameters in~\cite{maet}, these works rely on large synthetic datasets to achieve desired performance.~Unlike them, our FeatEnHancer is optimized from the task-related loss functions and does not require any pre-training on synthetic datasets mimicking low-light or harsh weather conditions.\\

\vspace{-10 pt}
\noindent\textbf{Other high-level vision tasks.} 
Besides face and object detection, recent research has explored high-level computer vision tasks like semantic segmentation~\cite{sem_seg_CRF_arxiv_14, sem_seg_FCN_CVPR_15}. Xue~\etal~\cite{see_understand_dark_22_ACMM} devise a contrastive-learning strategy to improve visual and machine perception simultaneously, achieving impressive performance on nighttime semantic segmentation of adverse conditions dataset with correspondences (ACDC) dataset~\cite{ACDC_dataset_ICCV_21}. Furthermore, DarkVision~\cite{DarkVision_dataset_arxiv_23} has emerged recently to tackle video object detection under low-light vision. In this work, thanks to~\cite {ACDC_dataset_ICCV_21, DarkVision_dataset_arxiv_23}, we apply FeatEnHancer to semantic segmentation and video object detection under low-light vision to investigate its generalization capabilities.

\subsection{Learning Multi-scale Hierarchical Features}
\label{sec:RW_MSHF}
Representing objects at varying scales is one of the main difficulties in computer vision.~Therefore, the work in this domain goes back to the era of hand-engineered features~\cite{lowe2004distinctive, dalal2005histograms, mikolajczyk2004scale, lazebnik2006beyond}. Modern object detectors~\cite{FasterRCNN, retinaNet, CascadeRCNN, deformable_detr, SparseRCNN, detectrs_cvpr_21, featurized-query-rcnn} exploit multi-scale features to tackle this challenge. Similarly, multi-scale representations~\cite{sem_seg_FCN_CVPR_15} and pyramid pooling schemes~\cite{Zhao_2017_CVPR} have been proposed for effective semantic segmentation. Moreover, current improvements in vision-based backbone networks~\cite{Res2Net_PAMI_2021, Swin_V2_CVPR22, Swin_ICCV_21} demonstrate that learning hierarchical features during feature extraction directly uplifts the downstream vision tasks~\cite{Mask_RCNN_ICCV_17, UpperNEt_ECCV_18, CascadeRCNN}. However, the multi-scale and hierarchical structures of CNN have not been fully explored for low-light vision tasks. 

Under harsh weather conditions, DENet~\cite{detection_driven_DENet} employs Laplacian Pyramid~\cite{burt1987laplacian} to decompose images into low and high-frequency components for object detection.~Despite the encouraging results, the multi-scale feature learning in DENet relies on the Laplacian pyramid, which is susceptible to noise and may produce inconsistencies in regions with high contrast or sharp edges. Alternatively, aligned with the multi-scale learning in modern vision backbone networks~\cite{FPN, Swin_ICCV_21, Swin_V2_CVPR22}, our FeatEnHancer employs CNN to generate multi-scale feature representations, which are fused through the scale-aware attentional feature aggregation and skip connections. Our approach is much more flexible and aligns with downstream vision tasks, boosting state-of-the-art results on multiple downstream vision tasks.

\section{Proposed Approach}
\label{sec:method}

The key idea of this paper is to design a general-purpose pluggable module that strengthens machine perception under low-light vision to solve several downstream vision tasks such as object detection, semantic segmentation, and video object detection. The overall architecture of FeatEnHancer is exhibited in Fig.~\ref{fig:featenhancer}. Our FeatEnHancer takes a low-light image as input and adaptively boosts its semantic representation by enriching task-related hierarchical features. We now discuss the key components of FeatEnHancer in detail.

\subsection{Hierarchical Feature Enhancement}
\label{sec:enhancing_hierarchical_features}

Inspired by the recent improvements in vision-based backbone networks~\cite{Res2Net_PAMI_2021, Swin_V2_CVPR22, Swin_ICCV_21}, we introduce the enhancement of hierarchical features through jointly optimizing feature enhancement and downstream tasks under low-light vision. Unlike~\cite{Res2Net_PAMI_2021, Swin_V2_CVPR22, Swin_ICCV_21}, our goal is to extract spatial features from low-light images and generate meaningful semantic representations.~In order to enhance hierarchical features, we first construct multi-scale representations from the low-light input image. Later, we feed these multi-scale representations to our feature enhancement network. \\

\noindent\textbf{Constructing multi-scale representations.}~~We take a low-light RGB image $I\in\mathbb{R}^{ H \times W \times C}$ as input and employ regular convolutional operator $\textbf{Conv}(.)$ on $I$ to generate $I_{q}\in\mathbb{R}^{ \frac{H}{4}\times\frac{W}{4} \times 3}$ and $I_{o}\in\mathbb{R}^{ \frac{H}{8}\times\frac{W}{8} \times 3}$ representing the \textit{quarter} and \textit{octa} scale of an input image, respectively. To summarize, it can be written as:
\begin{equation}
    \begin{aligned}
        I_{q} = \textbf{Conv}(I)~\qquad K = 7, S = 4, \\
        I_{o} = \textbf{Conv}(I_{q})\qquad K = 3, S = 2,
    \end{aligned}
    \label{eq:generating_scales}
\end{equation}
where $K$ and $S$ denote kernel size and stride, and $H$, $W$, and $C$ represent the height, width, and channels of an image.\\

\noindent\textbf{Feature enhancement network.}
In order to enhance features at each scale, we require an enhancement network that learns to enhance spatial information important for downstream tasks. Inspired by low-light image enhancement networks~\cite{zero-dce, zero-dce++}, we design a fully convolutional intra-scale feature extraction network (FEN). However, unlike~\cite{zero-dce, zero-dce++}, our FEN introduces a single convolutional layer at the beginning that generates a feature map $F\in\mathbb{R}^{ H \times W \times C}$, where $C$ is transformed from 3 to 32 by keeping the resolution ($H\times W$) same as the input. Then a series of six convolutional layers with symmetrical skip concatenation is applied, where each convolutional layer, with $K=3$ and $S=1$, is accompanied by the ReLU activation function.~We apply FEN on each scale $I$, $I_{q}$, and $I_{o}$ separately, and obtain multi-scale feature representations, denoted as $F$, $F_{q}$, and $F_{o}$, respectively.
This multi-scale learning allows the network to enhance both global and local information from higher and lower-resolution features. Hence, we ignore down-sampling and batch normalization to preserve semantic relations between neighbouring pixels which is similar to~\cite{zero-dce}. However, we discard the last convolutional layer of DCENet~\cite{zero-dce} in our FEN and propagate the final enhanced feature representations from each scale for the multi-scale feature fusion. Note that the implementation details of FEN in FeatEnHancer are independent of the proposed module, and even more, advanced image enhancement networks such as~\cite{STAR_ICCV} can be applied to improve performance. Now, we discuss multi-scale feature fusion in detail.

\subsection{Multi-scale Feature Fusion}
\label{sec:mult_scale_fusion}
Since we already have multi-scale feature representations ($F$, $F_{q}$, and $F_{o}$) from FEN, the remaining obstacle is to fuse them effectively.~Lower-scale features ($F_{o}$) contain fine details and edges.~In contrast, higher-resolution features ($F_{q}$) capture more abstract information, such as shapes and patterns.~Therefore, naive aggregation leads to inferior performance~(see Table~\ref{tab:effect_cross_attention}).~Hence,~we adopt two different strategies to capture both global and local information from higher and lower-resolution features.~First, inspired by multi-head attention in~\cite{attn_all_need_Neurips_17} that enables the network to jointly learn information from different channels, we design a scale-aware attentional feature aggregation (SAFA) module that jointly attends to features from different scales. Second, we adopt a skip connection~\cite{ResNet} (SC) scheme to integrate low-level information from $F_{o}$ and the enhanced representation from SAFA to obtain the final enhanced hierarchical representation. Adopting SAFA for merging high-resolution features and SC for lower-resolution features leads to a more robust hierarchical representation~(see Table~\ref{tab:multi_scale_combination}). Now, we discuss SAFA in detail.

\newcommand{\thickbar}[1]{\mathbf{\bar{\vphantom{#1}#1}}}
\noindent \textbf{Scale-aware Attentional Feature Aggregation.}~~Even though high-resolution features assist in capturing fine details, such as recognizing small objects, applying an attentional operation to them is computationally demanding. Thus, in SAFA, we propose an efficient multi-scale aggregation strategy where enhanced high-resolution hierarchical features are projected to a smaller resolution prior to attentional feature aggregation. As illustrated in Figure~\ref{fig:featenhancer}, SAFA transforms $F\in\mathbb{R}^{ H \times W \times C}$ to $Q\in\mathbb{R}^{ \frac{H}{8}\times\frac{W}{8} \times C}$ with two convolutional layers ($K=7, S=4; K=3, S=2$) and $F_{q}\in\mathbb{R}^{ \frac{H}{4}\times\frac{W}{4} \times C}$ to $K\in\mathbb{R}^{ \frac{H}{8}\times\frac{W}{8} \times C}$ with a single convolutional layer ($K=3, S=2$). Note that the weights of the convolutional layer ($K=3, S=2$) are not shared because, in addition to down-scaling the high-resolution features, it serves as an embedding network before computing the attentional weights.~Later, $Q$ and $K$ are concatenated to form the set of hierarchical features $F_{q+k}$, which are split into N blocks along the channel dimension $C$:
\begin{equation}
    \begin{aligned}
        F_{q+k}^{n} = F_{q+k}[:,:,(n-1)\frac{C}{N}:n\frac{C}{N}],
    \end{aligned}
    \label{eq:safa_split}
\end{equation}
where $n \in \{1,2,..., N\}$ and N is the total number of attentional blocks. The $F_{q+k}^{n}\in\mathbb{R}^{\frac{H}{8}\times\frac{W}{8} \times \frac{C}{N}}$ is used to compute attentional weights $W$ in a single attention block as follows:
\begin{equation}
    \begin{aligned}
        W_{q+k}^{n} = F_{q}^{n} \cdot F_{k}^{n},
    \end{aligned}
    \label{eq:safa_attention}
    \vspace{-10pt}
\end{equation}

\begin{equation}
    \begin{aligned}
        \thickbar{W}_{q+k}^{n} = \frac{\text{exp}(W_{q+k}^{n})}
        {\sum_{l=1}^L \text{exp}(W_{q+k}^{n})},
    \end{aligned}
    \label{eq:safa_softmax}
\end{equation}
where $W_{q+k}^{n}$ is the attentional weights of $F_{q}^{n}$ and $F_{k}^{n}$ for $n$-th block, and $\thickbar{W}_{q+k}^{n}$ is the normalized form of $W_{q+k}^{n}$. Derived from the $n$-th block of normalized attention weights, we apply weighted sum to compute the $n$-th block of enhanced hierarchical representation $\thickbar{F}_{h}^{n}\in\mathbb{R}^{\frac{H}{8}\times\frac{W}{8} \times \frac{C}{N}}$ as follows:
\begin{equation}
    \begin{aligned}
        \thickbar{F}_{h}^{n} = \sum_{l=1}^L\thickbar{W}_{q+k}^{n} \cdot F_{q+k}^{n},
    \end{aligned}
    \label{eq:safa_weighted_sum}
\end{equation}
now we concatenate all $\thickbar{F}_{h}^{n}$ along the channel dimension to obtain $\thickbar{F}_{h}\in\mathbb{R}^{\frac{H}{8}\times\frac{W}{8} \times C}$. Note that although $\thickbar{F}_{h}$ is the same size as $Q$ and $K$, it contains far richer representations, encompassing information from multi-scale high-resolution features.

Subsequently, as explained earlier in Sec.~\ref{sec:mult_scale_fusion}, 
with the help of skip connections (SC), we integrate $F_{o}$ and $\thickbar{F}_{h}$ to obtain the final enhanced hierarchical representation covering both global and local features, as illustrated in Figure~\ref{fig:intro_figure} and~\ref{fig:featenhancer}.~Note that prior to the skip connection, we upsample $\thickbar{F}_{h}$ and $F_{o}$, where the upsampling operation $U(.)\in\mathbb{R}^{\frac{H}{8}\times\frac{W}{8}\times C}\rightarrow\mathbb{R}^{H\times W \times C}$ is performed with simple bi-linear interpolation operation, which is much faster than using transposed convolutions~\cite{transpose_conv}. Unlike image enhancement in existing works~\cite{zero-dce, zero-dce++,maet}, with a multi-scale hierarchical feature enhancement strategy, our FeatEnHancer learns a powerful semantic representation by capturing both local and global features. This makes it a general-purpose module to enhance hierarchical features, boosting machine perception under low-light vision.
\begin{table}[ht]
\begin{adjustbox}{width=\linewidth}
\begin{tabular}{l|c|c|c|c}
   \textbf{Dataset} & \textbf{Task} & \textbf{\#Cls} & \textbf{\#Train} & \textbf{\#Val}\\
    \hline
    ExDark~\cite{exdark-dataset} & Dark object detection & 12 & 4800 & 2563 \\
    DARK FACE~\cite{DARK_FACE_dataset_1} & Face detection & 1 & 5400 & 600 \\
    ACDC Nightime~\cite{ACDC_dataset_ICCV_21} & Semantic segmentation & 19 & 400 & 106 \\
    DarkVision~\cite{DarkVision_dataset_arxiv_23} & Video object detection & 4 & 26 & 6 \\
    \bottomrule
\end{tabular}
\end{adjustbox}
\centering
\caption{\textbf{Statistics of the datasets} used to report results on four different downstream vision tasks. \#Cls is the number of classes, whereas \#Train and \#Val denote number of training and validation samples for each dataset, respectively.}.
\label{tab:DATASET}
\vspace{-20pt}
\end{table}

\definecolor{maroon}{cmyk}{0,0.87,0.68,0.32}

\section{Experiments}
\label{sec:experiments}
We conduct extensive experiments for evaluating the proposed FeatEnHancer module to several downstream tasks under the low-light vision, including generic object detection~\cite{exdark-dataset, nod}, face detection~\cite{DARK_FACE_dataset_1}, semantic segmentation~\cite{ACDC_dataset_ICCV_21}, and video object detection~\cite{DarkVision_dataset_arxiv_23}. Table~\ref{tab:DATASET} summarizes the crucial statistics of the employed datasets. This section first compares the proposed method with powerful baselines, existing LLIE approaches, and task-specific state-of-the-art methods. Then, we ablate the important design choices of our FeatEnHancer. We provide complete implementation details for each experiment in Appendix~\textcolor{red}{A}.

\begin{table*}
\begin{minipage}[t]{0.40\linewidth}
\centering
\footnotesize
\begin{adjustbox}{width=\linewidth}
 \begin{tabular}{c >{\columncolor{gray!20}}c c >{\columncolor{gray!20}}c c}
    \multirow{2}{*}{Methods} &
      \multicolumn{2}{c}{RetinaNet} &
      \multicolumn{2}{c}{FQ R-CNN}\\
      \cmidrule(lr){2-3}\cmidrule(lr){4-5}
      & { mAP$_{50}$} & {mAP} &  { mAP$_{50}$} & {mAP}\\
   \toprule
    Baseline & 72.1 & 46.3 & 74.5 & 47.0\\
    RAUS~\cite{raus} & 64.7 & 44.0 & 77.0  & 48.1 \\
    KIND~\cite{kind} & 70.7 & 45.1 & 80.5  & 51.5 \\
    Zero-DCE++~\cite{zero-dce++} & 70.3 & 45.2 & 79.5  & 49.2  \\
    EnGAN~\cite{enlightengan} & 70.4  & 44.9 & 80.0 & 51.9\\
    MBLLEN~\cite{mebbln}  & 70.6 & 45.1 & 80.0  & 51.0 \\
    Zero-DCE~\cite{zero-dce} & 71.0 & 45.2 & 80.6  & 52.0 \\
    MAET~\cite{maet} & 71.8 & 45.7 & 81.6 & 52.4\\
    \textbf{FeatEnHancer} & \textbf{72.6} &\textbf{ 46.4} & \textbf{86.3}  & \textbf{56.5}\\
    \bottomrule
  \end{tabular}
  \end{adjustbox}
 \caption{\textbf{Quantitative comparison on ExDark dataset.} Results obtained on the commonly used evaluation metrics are highlighted. Our FeatEnHancer brings consistent improvements and achieves new state-of-the-art results with FQ R-CNN.}
 \label{tab:results_exdark}
\end{minipage}
\hfill
\begin{minipage}[t]{0.37\linewidth}
\centering
\footnotesize
\begin{adjustbox}{width=\linewidth}
 \begin{tabular}{c >{\columncolor{gray!20}}c c >{\columncolor{gray!20}}c c}
    \multirow{2}{*}{Methods} &
      \multicolumn{2}{c}{RetinaNet} &
      \multicolumn{2}{c}{FQ R-CNN}\\
      \cmidrule(lr){2-3}\cmidrule(lr){4-5}
      & { AP$_{50}$} & {AP} &  { AP$_{50}$} & {AP}\\
   \toprule
    Baseline & 47.3  & 19.9 & 67.5 & 28.6\\
    RAUS~\cite{raus} & 42.1  & 17.6 & 65.5 & 27.4 \\
    KIND~\cite{kind} & 47.2 & 19.8 & 65.0  & 27.5  \\
    Zero-DCE++~\cite{zero-dce++} & 47.3  & 20.1 & 66.2 & 28.2  \\
    EnGAN~\cite{enlightengan} & 45.1 & 19.3 & 67.4  & 28.4\\
    MBLLEN~\cite{mebbln}  & 47.1 & 19.8 & 67.3 & 27.1 \\
    Zero-DCE~\cite{zero-dce}  & \textbf{47.4} & \textbf{20.1} & 66.9  & 27.5 \\
    MAET~\cite{maet} & 44.3 & 18.7 & 66.1 & 27.1\\
    \textbf{FeatEnHancer} & 47.2 & 19.9 & \textbf{69.0}  & \textbf{29.4}\\
    \bottomrule
  \end{tabular}
\end{adjustbox}
\caption{\textbf{Comparing FeatEnHancer on the DARK FACE dataset}. With RetinaNet, FeatEnHancer performs on par with other methods. However, with FQ R-CNN, FeatEnHancer surpasses all of them.}
\label{tab:results_dark_face}
\end{minipage}
\hfill
\begin{minipage}[t]{0.195\linewidth}
\begin{adjustbox}{width=\linewidth}
\centering
  \begin{tabular}{c c} 
    Method & mIoU \\
    \hline
    Baseline~\cite{deeplabv3_ECCV_2018} & 45.7 \\
    RetinexNet~\cite{retinex_bmvc_18} & 41.9 \\
    DRBN~\cite{DRBN} & 43.3 \\
    FIDE~\cite{FIDE} & 43.4 \\
    KIND~\cite{kind} & 43.0 \\
    EnGAN~\cite{enlightengan} & 43.8 \\
    ZeroDCE~\cite{zero-dce} & 43.4 \\
    SSIENet~\cite{ssie_net_arxiv_20} & 41.4 \\
    Xue~\etal\cite{see_understand_dark_22_ACMM} & 49.8 \\
     \rowcolor{gray!25} \textbf{FeatEnHancer} &  \textbf{54.9} \\
\bottomrule
 \end{tabular}  
 \end{adjustbox}
 \caption{\textbf{Quantitative comparison on the ACDC dataset.} Huge gains from our FeatEnHancer lead to new state-of-the-art results.}
 \label{tab:results_acdc}
\end{minipage}
\vspace{-10pt}
\end{table*}

\subsection{Dark Object Detection}
\label{sec:generic_dark_object_detection}

\noindent\textbf{Settings.}~~For dark object detection experiments on the real-world data, we consider the exclusively dark (ExDark)~\cite{exdark-dataset} dataset (see Table~\ref{tab:DATASET}). We adopt RetinaNet~\cite{retinaNet} as a typical detector and Featurized Query R-CNN~\cite{featurized-query-rcnn} (FQ R-CNN) as an advanced object detection framework to report results. In the case of both detectors, pre-trained models on COCO~\cite{coco-dataset} are fine-tuned on each dataset. For RetinaNet, images are resized to 640$\times$640, and we train the network using 1$\times$schedule in mmdetection~\cite{mmdetection} (12 epochs using SGD optimizer~\cite{sgd_optimizer} with an initial learning rate of 0.001). For Featurized Query R-CNN, we employ multi-scale training~\cite{DETR, SparseRCNN, featurized-query-rcnn} (shorter side ranging from 400 to 800 with a longer side of 1333). The FQ R-CNN is trained for 50000 iterations using ADAMW~\cite{adamw_optimizer} optimizer (initial learning rate of 0.0000025, weight decay of 0.0001, and batch size of 8). Note that for each object detection framework, we adopt the same settings while reproducing results of our work, baseline, LLIE approaches, and task-specific state-of-the-art methods.

We compare our FeatEnHancer to several state-of-the-art LLIE methods, including KIND~\cite{kind}, RAUS~\cite{raus}, EnGAN~\cite{enlightengan}, MBLLEN~\cite{mebbln}, Zero-DCE~\cite{zero-dce}, Zero-DCE++~\cite{zero-dce}, and state-of-the-art dark object detection method, MAET~\cite{maet}. For LLIE methods, all images are enhanced from their released checkpoints and propagated to the detector.~In case of MAET~\cite{maet}, we pre-train the detector using their proposed degrading pipeline and then fine-tune it on both datasets to establish a direct comparison.\\

\begin{figure}
\centering
\includegraphics[width=\linewidth]{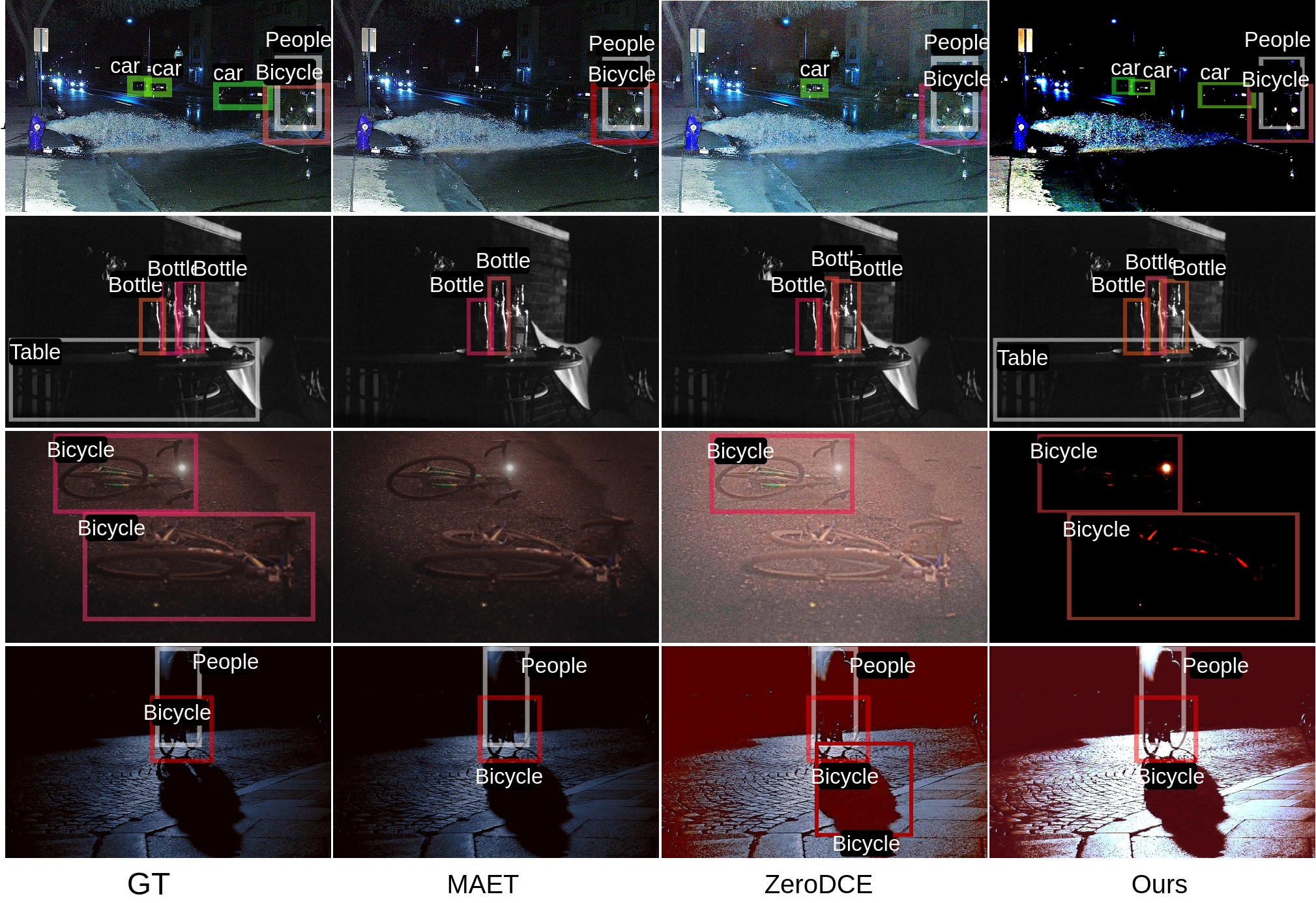}
   \caption{\textbf{Visual comparison of FeatEnHancer with the two previous best competitors on the ExDark dataset}. Zoom in for the best view.}
\label{fig:qual_exdark}
\vspace{-10pt}
\end{figure}

\noindent\textbf{Results on ExDark.}~~Table~\ref{tab:results_exdark} lists the results of LLIE works, MAET, and the proposed method on both object detection frameworks.~It is evident that our FeatEnHancer brings consistent and significant gains over prior methods. Note that, while the performance of MAET and our method is comparable on RetinaNet~($\approx$ 72 AP$_{50}$), the proposed FeatEnHancer outperforms MAET by a significant margin on FQ R-CNN, achieving the new state-of-the-art AP$_{50}$ of 86.3. Furthermore, Figure~\ref{fig:qual_exdark} shows four detection examples from our method and the two best competitors using FQ R-CNN as a detector. These results illustrate that despite the inferior visual quality, our FeatEnHancer enhances hierarchical features that are favourable for dark object detection, producing state-of-the-art results.


\subsection{Face Detection on DARK FACE}
\label{sec:face_detection}
\noindent\textbf{Settings.}~~The DARK FACE~\cite{DARKFACE_dataset_2, DARK_FACE_dataset_1} is a challenging face detection dataset released for the UG\textsuperscript{2} competition.~For experiments on the DARK FACE (see Table~\ref{tab:DATASET}), the images are resized to a larger resolution of $1500\times1000$ for all methods. We adopt the same object detection frameworks of RetinaNet and FQ R-CNN and follow identical experimental settings, as explained in Sec.~\ref{sec:generic_dark_object_detection}.



\noindent\textbf{Results.}~~The performance of FeatEnHancer, MAET, and six LLIE methods, using RetinaNet and Featurized Query R-CNN, are summarized in Table~\ref{tab:results_dark_face}. Note that a few LLIE methods~\cite{zero-dce, zero-dce++, kind} yield superior results than our approach in the case of RetinaNet. We argue that 
due to tiny faces with highly dark images in the DARK FACE dataset, RetinaNet fails to capture information even from the enhanced hierarchical features. We discuss this behaviour with an example in Appendix~\textcolor{red}{B}. On the other hand, LLIE approaches directly provides well-lit images that bring slightly bigger gains (+0.1 mAP$_{50}$) in this case. However, note that with the more strong detector, our FeatEnHancer surpasses all the LLIE methods and MAET by a significant margin (+1.5 mAP$_{50}$), achieving mAP$_{50}$ of 69.0.



\subsection{Nighttime Semantic Segmentation on ACDC}
\label{sec:semseg_acdc}
\noindent\textbf{Settings.}~~We utilize nighttime images from the ACDC dataset~\cite{ACDC_dataset_ICCV_21} (see Table~\ref{tab:DATASET}) to report results on semantic segmentation in a low-light setting.~DeepLabV3+~\cite{deeplabv3_ECCV_2018} is adopted as the segmentation baseline from mmseg~\cite{mmseg2020} for straightforward comparison with the concurrent work~\cite{see_understand_dark_22_ACMM}. We follow identical experimental settings as in~\cite{see_understand_dark_22_ACMM}.~Refer to Appendix~\textcolor{red}{{A}} for complete implementation details.\\

\begin{figure}
\centering
\includegraphics[width=\linewidth]{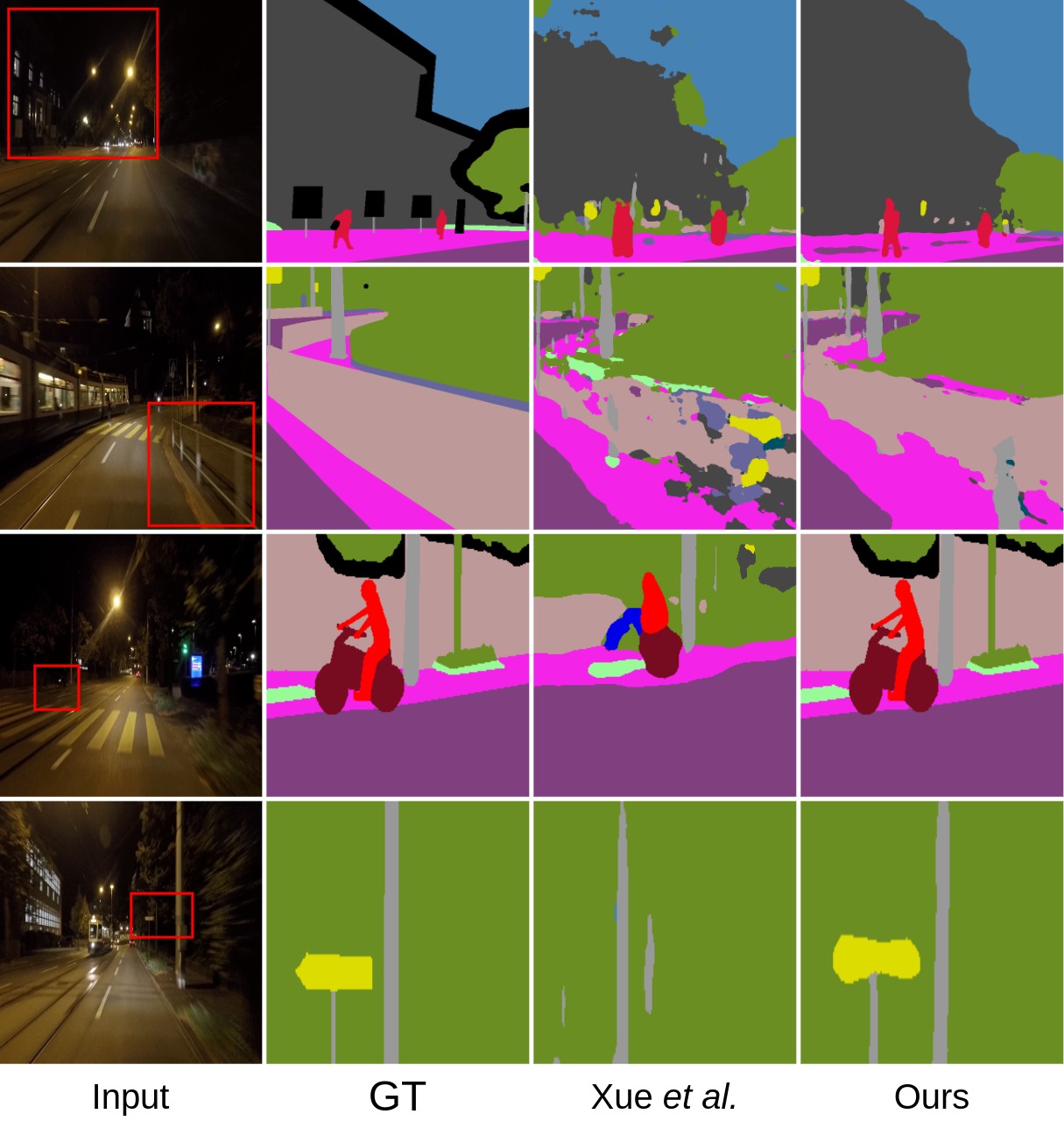}
    \caption{\textbf{Qualitative comparison of FeatEnHancer with previous best work~\cite{see_understand_dark_22_ACMM} on the ACDC nighttime semantic segmentation.} FeatEnHancer provides more accurate segmentations.}
\label{fig:qual_ACDC}
\vspace{-10pt}
\end{figure}

\begin{table}[ht]
\centering
\small
\begin{tabular}{ccc}
    \multirow{2}{*}{Method } & Illumination (3.2) & Illumination (0.2)\\
    & mAP &mAP \\
    \toprule
    Baseline~\cite{SELSA_ICCV_19} & 32.8  & 10.4 \\
    RAUS~\cite{raus}  & 7.42 & 5.19 \\
    EnGAN~\cite{enlightengan}   & 7.83& 5.41 \\
    MBLLEN~\cite{mebbln}  & 7.82 & 5.39 \\
    KIND~\cite{kind}  & 7.43& 5.25 \\
    Zero-DCE++~\cite{zero-dce++}  & 7.51& 5.02 \\
    Zero-DCE~\cite{zero-dce}  & 7.83 & 5.43\\
    \rowcolor{gray!20}\textbf{FeatEnHancer} & \textbf{34.6} & \textbf{11.2} \\
    \bottomrule
\end{tabular}
\centering
\caption{\textbf{Comparing FeatEnHancer with LLIE methods on the DarkVision dataset}. FeatEnHancer is the only method that boosts the performance of the powerful baseline method on both illumination levels.}.
\label{tab:results_dakrvision}
\vspace{-20pt}
\end{table}

\noindent\textbf{Results. } We compare our method with several state-of-the-art LLIE methods, including RetinexNet~\cite{retinex_bmvc_18} KIND~\cite{kind}, FIDE~\cite{FIDE}, DRBN~\cite{DRBN}, EnGAN~\cite{enlightengan}, SSIENet~\cite{ssie_net_arxiv_20}, ZeroDCE~\cite{zero-dce}, and current state-of-the-art nighttime semantic segmentation method Xue~\etal~\cite{see_understand_dark_22_ACMM}. As shown in Table~\ref{tab:results_acdc}, our FeatEnHancer brings remarkable improvements in the baseline with a mIoU of 54.9, outperforming the previous best result by 5.1 points. Moreover, we present a qualitative comparison with the previous best competitor~\cite{see_understand_dark_22_ACMM} in Figure~\ref{fig:qual_ACDC}. Evidently, our FeatEnHancer generates more accurate segmentation for both bigger and smaller objects, such as terrain and traffic signs in the last row. These results affirm the effectiveness of FeatEnHancer as a general-purpose module achieving state-of-the-art results in both dark object detection and nighttime semantic segmentation.

\subsection{Video Object Detection on DarkVision}
\label{sec:vod_darkvision}
\noindent\textbf{Settings.}~~We extend our experiments from static images to video domain to test the generalization capabilities of our method. The video object detection under low-light vision is evaluated on the recently emerged~DarkVision~dataset~\cite{DarkVision_dataset_arxiv_23} (see Table~\ref{tab:DATASET} for dataset details). Although the dataset is not publicly available yet, we sincerely thank the authors of~\cite{DarkVision_dataset_arxiv_23} for providing prompt access. To evaluate our FeatEnHancer under low light settings, we take the low-end camera split on two different illumination levels, i.e., 0.2 and 3.2. For ablation studies, we adopt a 3.2\% illumination level split.~We consider SELSA~\cite{SELSA_ICCV_19} as our baseline and follow identical experimental settings with the ResNet-50 backbone network in the mmtracking~\cite{mmtrack2020}. To establish a direct comparison, we enhance all video frames first through LLIE methods and feed these frames to the baseline, as done in Sec.~\ref{sec:generic_dark_object_detection}. As a common practice in video object detection~\cite{temporal_roi_align_AAAI_21, sparsevod_bmvc_22, SELSA_ICCV_19}, the mAP@IoU=0.5 is utilized as an evaluation metric to report results. More details can be found in Appendix~\textcolor{red}{{A}}.\\

\noindent\textbf{Results.}~~Table~\ref{tab:results_dakrvision} compares our FeatEnHancer with several LLIE methods~\cite{raus, enlightengan, mebbln, kind, zero-dce, zero-dce++} and the powerful video object detection baseline~\cite{SELSA_ICCV_19}. Evidently, our FeatEnHancer provides considerable gains to the baseline with 34.6 mAP and 11.2 mAP under illumination levels of 3.2 and 0.2, respectively. Note that our FeatEnHancer is the only method that boosts performance under both image and video modalities. In contrast, as shown in Table~\ref{tab:results_dakrvision}, existing LLIE methods not only fail to assist the baseline method but also deteriorate the performance. This poor generalization of LLIE approaches highlights that learning from domain-specific paired data~\cite{mebbln, kind, raus}, unpaired data~\cite{enlightengan}, and curve estimation without data~\cite{zero-dce, zero-dce++} are not the optimal solutions for generalized enhancement methods. Hence, more research is required.


\definecolor{maroon}{cmyk}{0,0.87,0.68,0.32}

\begin{table*}[ht]
\centering
\begin{subtable}{0.4\linewidth}
\begin{adjustbox}{width=\linewidth}
\begin{tabular}{c|c|c|c}
    \multirow{2}{*}{Method } & ExDark & ACDC & DarkVision \\
    & (mAP) & (mIoU) &(mAP) \\
    \hline
    simple averaging & 69.5 & 50.3 & 32.9 \\
    skip connections~\cite{ResNet} & 70.3& 51.7 & 33.1 \\
    \rowcolor{gray!20}SAFA & 72.6 & 54.9 & 34.6 \\
    \bottomrule
\end{tabular}
\end{adjustbox}
\centering
\caption{\textbf{Effectiveness of SAFA.}}
\label{tab:effect_cross_attention}
\end{subtable}
\hspace{20pt}
\begin{subtable}{0.35\linewidth}
\begin{adjustbox}{width=\linewidth}
\begin{tabular}{c|c|c|c}
    \multirow{2}{*}{Method } & ExDark & ACDC & DarkVision \\
    & (mAP) & (mIoU) &(mAP) \\
    \hline
    SC , SC & 69.7& 51.7 & 32.8 \\
    SAFA , SAFA & 70.2& 52.6 &33.4 \\
    SC, SAFA & 70.9 & 52.9 & 33.8 \\
    \rowcolor{gray!20} SAFA, SC & 72.6 & 54.9 & 34.6 \\
    \bottomrule
\end{tabular}
\end{adjustbox}
\centering
\caption{\textbf{Various combinations of multi-scale fusion.}}
\label{tab:multi_scale_combination}
\end{subtable}\\
\begin{subtable}{0.37\linewidth}
\begin{adjustbox}{width=\linewidth}
\begin{tabular}{c|c|c|c}
    \multirow{2}{*}{Method} & ExDark & ACDC & DarkVision \\
    & (mAP) & (mIoU) &(mAP) \\
    \hline
    maxpool & 69.3 & 51.3 & 32.9 \\
    adavgpool~\cite{STAR_ICCV} & 69.9 & 50.7 & 32.9 \\
    interpolation~\cite{image_adaptive_YOLO_AAAI_22} & 70.7 & 51.5 & 33.1 \\
    \rowcolor{gray!20} Convolution & 72.6 & 54.9 & 34.6 \\
    \bottomrule
\end{tabular}
\end{adjustbox}
\centering
\caption{\textbf{Downsampling approaches.}}
\label{tab:downsampling_ablation}
\end{subtable}
\begin{subtable}{0.3\linewidth}
\begin{adjustbox}{width=\linewidth}
\begin{tabular}{c|c|c|c}
    \multirow{2}{*}{Scale} & ExDark & ACDC & DarkVision \\
    & (mAP) & (mIoU) &(mAP) \\
    \hline
    (2, 4)& 71.8 & 52.7 & 34.1 \\
    \rowcolor{gray!20}(4, 8)& 72.6 & 54.9 & 34.6 \\
    (4, 16)& 71.5 & 51.4 & 33.9 \\
    (8, 16)& 68.7 & 45.6 & 31.9 \\
    \bottomrule
\end{tabular}
\end{adjustbox}
\centering
\caption{\textbf{Scales for $I_{q}$ and $I_{o}$, respectively.}}
\label{tab:scale_selection}
\end{subtable}
\begin{subtable}{0.27\linewidth}
\begin{adjustbox}{width=\linewidth}
\begin{tabular}{c|c|c|c}
    \multirow{2}{*}{$N$ } & ExDark & ACDC & DarkVision \\
    & (mAP) & (mIoU) &(mAP) \\
    \hline
    2 & 72.1 & 53.9 & 34.2 \\
    4 & 72.4 & 54.3 & 34.5 \\
    \rowcolor{gray!20}8 & 72.6 & 54.9 & 34.6 \\
    12 & 72.4 & 54.3 & 34.1 \\
    \bottomrule
\end{tabular}
\end{adjustbox}
\centering
\caption{\textbf{ \# attentional blocks in SAFA.}}
\label{tab:attention_blocks}
\end{subtable}
\caption{\textbf{Ablations for the proposed FeatEnHancer} on three benchmarks. \textbf{(a)} We investigate the effectiveness of SAFA by replacing it with different aggregation methods to fuse $F$ and $F_{q}$. \textbf{(b)} We experiment with various combinations of SAFA and skip connection (SC) to justify an optimal design choice. Here, (SC, SC) means employing only skip connections to merge both $F_{q}$ and $F_{o}$ with $F$. \textbf{(c)} Besides convolution, we experiment with other downsampling techniques to generate lower resolutions. Here, adavgpool denotes adaptive average pooling as done in~\cite{STAR_ICCV}. \textbf{(d)} We vary scale sizes to generate lower-scale representations. Here, (2, 4) means $I_{q}\in\mathbb{R}^{\frac{H}{2}\times \frac{W}{2} \times 3}$ and $I_{o}\in\mathbb{R}^{\frac{H}{4}\times \frac{W}{4} \times 3}$. \textbf{(e)} We vary number of attentional blocks $N$ in SAFA of FeatEnHancer. Default settings are \colorbox{gray!20}{highlighted}.}
\label{tab:example}
\vspace{-5pt}
\end{table*}

\subsection{Ablation Studies}
\label{sec:ablation_studies}

This section ablates important design choices in the proposed FeatEnHancer when plugged into RetinaNet, DeeplabV3+, and SELSA on ExDark (dark object detection), ACDC (nighttime semantic segmentation), and DarkVision with illumination level of 3.2\% (video object detection), respectively.

\noindent\textbf{SAFA in FeatEnHancer.}~~The important component of the proposed FeatEnHancer is the scale-aware attentional feature aggregation~(SAFA) that aggregates high-resolution features. To validate its effectiveness, we conduct multiple experiments where SAFA is replaced with simple averaging or skip connections (SC)~\cite{ResNet} to fuse enhanced multi-scale features $F$ and $F_{q}$ (see Sec.~\ref{sec:mult_scale_fusion}). The experiment results are summarized in Table~\ref{tab:effect_cross_attention}. It is clear that SAFA outperforms both averaging and SC strategies by +2.3 mAP on ExDark, +3.2 mIoU on ACDC, and +1.5 mAP on DarkVision. These significant boosts across all three benchmarks indicate that scale-aware attention leads to optimal multi-scale feature aggregation in the proposed FeatEnHancer.

\noindent\textbf{Multi-scale feature fusion.}~~We experiment with various combinations of SAFA and SC to find an optimal design choice to fuse $F_q$ and $F_o$ with $F$ (see Sec.~\ref{sec:mult_scale_fusion}). As shown in Table~\ref{tab:multi_scale_combination}, there is a clear increase in performance, achieving (72.6 mAP on ExDark, 54.9 mIoU on ACDC, and 34.6 mAP on DarkVision) when SAFA is applied to fuse $F$ and $F_q$ first, and then $F_o$ is merged with the output of SAFA using skip connection. Hence, we use this approach as the default setting.

\noindent\textbf{Convolutional Downsampling.}~~Table~\ref{tab:downsampling_ablation} summarizes results from different downsampling techniques applied on the input image $I$ to generate lower-resolutions $I_q$ and $I_o$ (see Sec~\ref{sec:enhancing_hierarchical_features}). Our proposed convolutional downsampling yields impressive gains of +1.9 mAP on ExDark, +3.4 mIoU, and +1.5 mAP on DarkVision compared to max-pooling, adaptive average pooling ~\cite{STAR_ICCV}, and bilinear interpolation ~\cite{image_adaptive_YOLO_AAAI_22}. These results demonstrate the effectiveness of convolutional downsampling since it is better aligned with various vision backbone networks~\cite{Swin_ICCV_21, Res2Net_PAMI_2021, FPN}.

\noindent\textbf{Different Scale sizes.}~~We analyse the effect of different scale sizes to generate lower resolutions in Table~\ref{tab:scale_selection}. Here, for instance, (2, 4) means that the resolution of input image $I\in\mathbb{R}^{H\times W\times 3}$ is reduced by a factor of 2 and 4 to generate $I_{q}\in\mathbb{R}^{\frac{H}{2}\times \frac{W}{2} \times 3}$ and $I_{o}\in\mathbb{R}^{\frac{H}{4}\times \frac{W}{4} \times 3}$, respectively. Note that all these scales are generated through regular convolutional operator $\textbf{Conv}(.)$, as explained in Eq.~\ref{eq:generating_scales}. Looking at results in Table~\ref{tab:scale_selection}, the top performance on all three tasks is achieved with the scale size of (4, 8), thereby, preferred as a default setting.

\noindent\textbf{Number of Attention Blocks in SAFA.}~~Table~\ref{tab:attention_blocks} studies the effect of the number of attention blocks $N$ in our SAFA. The performance rises for all three tasks with the increase in N. This demonstrates that more attentional blocks in SAFA bring additional gains. The best performance with 72.6 mAP on ExDark, 54.9 mIoU on ACDC, and 34.6 mAP on DarkVision is achieved when $N$ reaches 8, and after that, it tends to saturate. Hence, $N=8$ is used as the default setting.



\section{Conclusion}
\label{sec:conclusion}
This paper proposes FeatEnHancer, a novel general-purpose feature enhancement module designed to enrich hierarchical features favourable for downstream tasks under low-light vision. Our intra-scale feature enhancement and scale-aware attentional feature aggregation schemes are aligned with vision backbone networks and produce powerful semantic representations.~Furthermore, our FeatEnHancer neither requires pre-training on synthetic datasets nor relies on enhancement loss functions. These architectural innovations make FeatEnHancer a plug-and-play module. Extensive experiments on four different downstream visions tasks covering both images and videos demonstrate that our method brings consistent and significant improvements over baselines, LLIE methods, and task-specific state-of-the-art approaches.

\noindent\textbf{Ethical Considerations.}~~Our work focuses on developing innovative strategies to learn optimal features from less illuminated real-world data to solve high-level vision tasks. Although the proposed method seems unbiased from an ethical perspective, as with any trained model, we recommend a thorough validation prior to deployment.





{\small
\bibliographystyle{ieee_fullname}
\bibliography{egbib}
}

\clearpage

\renewcommand{\thesection}{\Alph{section}}
\renewcommand{\thesubsection}{\thesection\arabic{subsection}}
\setcounter{section}{0}

\noindent\textbf{Appendix}\\ 

\noindent\textbf{Overview:}~ We first provide complete implementation details of our experiments on dark object detection in Appendix~\ref{sec:dark_obj}, face detection in Appendix~\ref{sec:face_det}, semantic segmentation in Appendix~\ref{sec:sem_seg}, and video object detection in Appendix~\ref{sec:vod}. Then, we present the quantitative and qualitative analysis in Appendix~\ref{sec:results_exdark},~\ref{sec:results_darkface}, and~\ref{sec:results_ACDC}. We compare and visualize the learned features between RetinaNet and Featurized Query R-CNN in Appendix~\ref{sec:retina_fqrcnn}. We discuss the performance in terms of learned representation and predictions between our intra-scale feature enhancement network and the DCENet~\cite{zero-dce, zero-dce++} in Appendix~\ref{sec:fen_against_dce}. The performance comparison between our FeatEnHancer and unsupervised domain adaptation methods is provided in Appendix~\ref{sec:domain_adaptation}. Finally, in Appendix~\ref{sec:comp_efficiency}, we discuss the performance-efficiency trade-off between our FeatEnHancer, Low-Light Image Enhancement (LLIE) approaches, and task-specific state-of-the-art methods.

\renewcommand{\thefigure}{\Roman{figure}} 
\renewcommand{\thetable}{\Roman{table}} 

\section{Implementation Details}
\label{sec:imp_details}

\subsection{Dark Object Detection}
\label{sec:dark_obj}

For dark object detection experiments on real-world data, we adopt RetinaNet~\cite{retinaNet} as a typical detector and Featurized Query R-CNN~\cite{featurized-query-rcnn} (FQ R-CNN) as an advanced object detection framework to report results. The implementation of FeatEnHancer with FQ R-CNN is based on detectron2~\cite{detectron2} with ResNet-101~\cite{ResNet} as the backbone network pre-trained with COCO~\cite{coco-dataset} weights. For the training, a batch size of 8 is employed with all the images resized to a maximum scale of 1333 $\times$ 800. Our training follows a $50K$ scheduler using ADAMW~\cite{adamw_optimizer} optimizer with an initial learning rate set to 0.0000025, which is divided by 10 both at 42000 and 47000 iterations. All experiments are carried out on RTX-2080 Ti GPU.

For RetinaNet, images are resized to 640$\times$640, and we train the network using 1$\times$schedule\footnote{\url{https://github.com/open-mmlab/mmdetection/blob/master/configs/_base_/schedules/schedule_1x.py}} in mmdetection~\cite{mmdetection} (12 epochs using SGD optimizer~\cite{sgd_optimizer} with an initial learning rate of 0.001). For evaluation, along with the common practice of employing mAP@IoU=0.5, we report mAP@IoU=0.5:95 using~\cite{coco-dataset} for completeness. Note that for each object detection framework, we adopt the same settings while reproducing results of our work, baseline, LLIE approaches, and task-specific state-of-the-art methods.

We compare our FeatEnHancer to several state-of-the-art LLIE methods, including KIND~\cite{kind}, RAUS~\cite{raus}, EnGAN~\cite{enlightengan}, MBLLEN~\cite{mebbln}, Zero-DCE~\cite{zero-dce}, Zero-DCE++~\cite{zero-dce}, and state-of-the-art dark object detection method, MAET~\cite{maet}. For LLIE methods, all images are enhanced from their released checkpoints and propagated to the detector.~In case of MAET~\cite{maet}, we pre-train the detector using their proposed degrading pipeline and then fine-tune it on both datasets to establish a direct comparison.

\subsection{Face Detection}
\label{sec:face_det}

The DARK FACE~\cite{DARK_FACE_dataset_1} dataset comprises dark human faces of various sizes in low-illuminated environments. In order to capture the tiny human faces, the images are resized to a maximum of 1333 $\times$ 800 and 1500 $\times$ 1000 for FQ R-CNN~\cite{featurized-query-rcnn} and RetinaNet~\cite{retinaNet}, respectively. The other settings and hyperparameters are identical to the Dark Object detection experiments explained in Appendix.~\ref{sec:dark_obj}. 

\subsection{Semantic Segmentation}
\label{sec:sem_seg}

The semantic segmentation with FeatEnHancer is implemented using MMSegmentation~\cite{mmsegmentation}, where the images are resized to 2048 $\times$ 1024 for the training. For direct comparison with previous state-of-the-art method~\cite{see_understand_dark_22_ACMM}, we use DeepLabV3+~\cite{deeplabv3_ECCV_2018}, an encoder-decoder style segmentor with ResNet-50~\cite{ResNet} as the backbone for nighttime semantic segmentation. The backbone is initialized with pre-trained ImageNet~\cite{imagenet_image_cvpr_09} weights, and we use a batch size of 4 for the training. The SGD~\cite{sgd_optimizer} optimizer following a $20K$ scheduler\footnote{\url{https://github.com/open-mmlab/mmsegmentation/blob/master/configs/_base_/schedules/schedule_20k.py}} of MMSegmentation~\cite{mmsegmentation} is employed with a base learning rate of 0.01 and a weight decay of 0.0005. For the direct comparison with the LLIE methods, we enhance all the images before passing them to the segmentor as done in Appendix.~\ref{sec:dark_obj}. The Mean Intersection over Inion (mIoU) is used to report the segmentation results in comparison to the baseline, LLIE approaches and existing state-of-the-art method Xue~\etal\cite{see_understand_dark_22_ACMM}.

\subsection{Video Object Detection}
\label{sec:vod}
Besides object detection and semantic segmentation on images, we extend our experiments to the video domain to test the generalization capabilities of the proposed FeatEnHancer. The video object detection under low-light vision is evaluated on the recently emerged~DarkVision~dataset~\cite{DarkVision_dataset_arxiv_23} (see Table~\textcolor{red}{1} in the main paper). Although the dataset is not publicly available yet, we sincerely thank the authors of~\cite{DarkVision_dataset_arxiv_23} for providing prompt access to its subset. To evaluate our FeatEnHancer under the low-light setting, we take the low-end camera split on two different illumination levels, i.e., 0.2 and 3.2. For ablation studies, we adopt a 3.2\% illumination level split.

We consider SELSA~\cite{SELSA_ICCV_19} as our baseline due to its simple and effective design and impressive performance on video object detection benchmarks~\cite{russakovsky2015imagenet, epic_kitchen_eccv_18}. As the backbone network, we use ResNet-50~\cite{ResNet} pre-trained on ImageNet~\cite{imagenet_image_cvpr_09}. For the detection, we apply Region Proposal Network (RPN)~\cite{FasterRCNN} on the output of \textit{conv4} to generate candidate proposals on each frame. In RPN, a total of 12 anchors with four scales $\{4, 8, 16, 32\}$ and three aspect ratios $\{0.5, 1.0, 2.0\}$ are used. The final 300 proposals are selected from each frame. In summary, we follow identical experimental settings by following the config of 1$\times$ schedule\footnote{\url{https://github.com/open-mmlab/mmtracking/blob/master/configs/vid/selsa/selsa_faster_rcnn_r50_dc5_1x_imagenetvid.py}} in mmtracking~\cite{mmtrack2020}. 

To establish a direct comparison, we enhance all video frames first through LLIE methods and feed these frames to the baseline, as done in Appendix.~\ref{sec:dark_obj}. For our method, we integrate FeatEnHancer in the baseline in an end-to-end fashion (see Fig.~\textcolor{red}{2} in the main paper).  Following standard practice in video object detection~\cite{temporal_roi_align_AAAI_21, sparsevod_bmvc_22, SELSA_ICCV_19}, the mAP@IoU=0.5 is utilized as an evaluation metric to report results. \textit{Note that the goal of this experiment is not to surpass prior state-of-the-art results on dark video object detection~\cite{DarkVision_dataset_arxiv_23}. Instead, the target is to demonstrate the effectiveness and generalization capabilities of the proposed FeatEnHancer in the video domain. Furthermore, we believe that far better results can be attained by incorporating our FeatEnHancer with better baselines and optimal experimental configurations.}

\begin{figure*}
\begin{center}
\includegraphics[width=\linewidth]{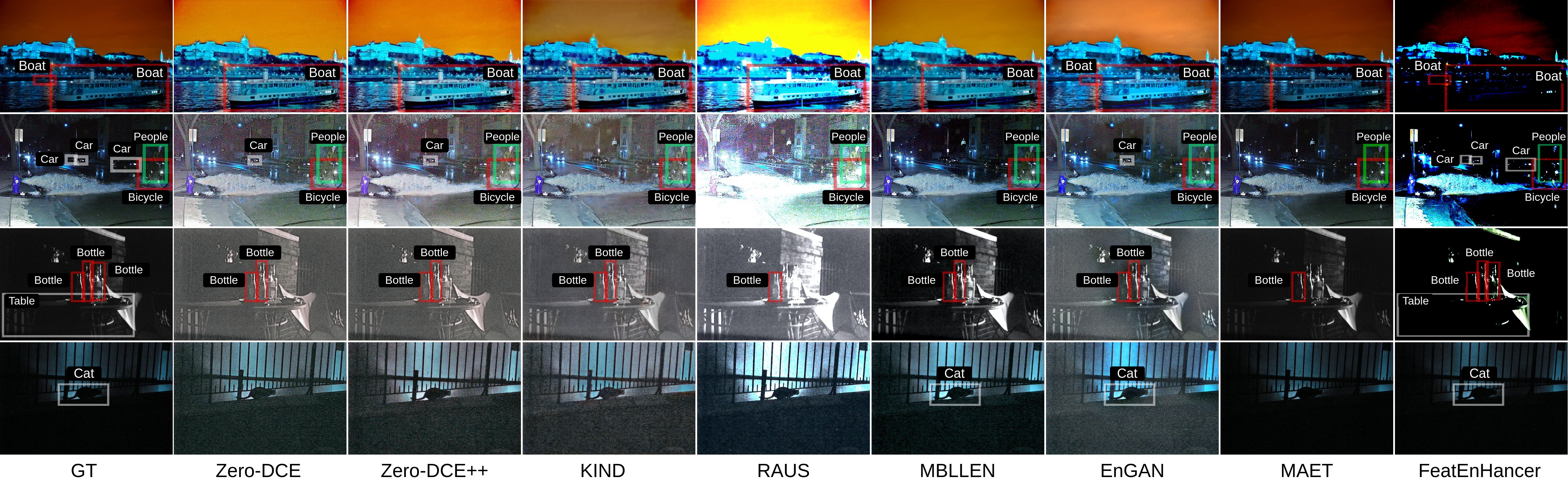}
   \caption{\textbf{Visual comparison of FeatEnHancer with several LLIE approaches and a previous state-of-art dark object detection method on the ExDark dataset}. Zoom in for the best view.}
\label{fig:qual_exdark_bigger}
\end{center}
\end{figure*}
\begin{figure*}
\begin{center}
\includegraphics[width=\linewidth]{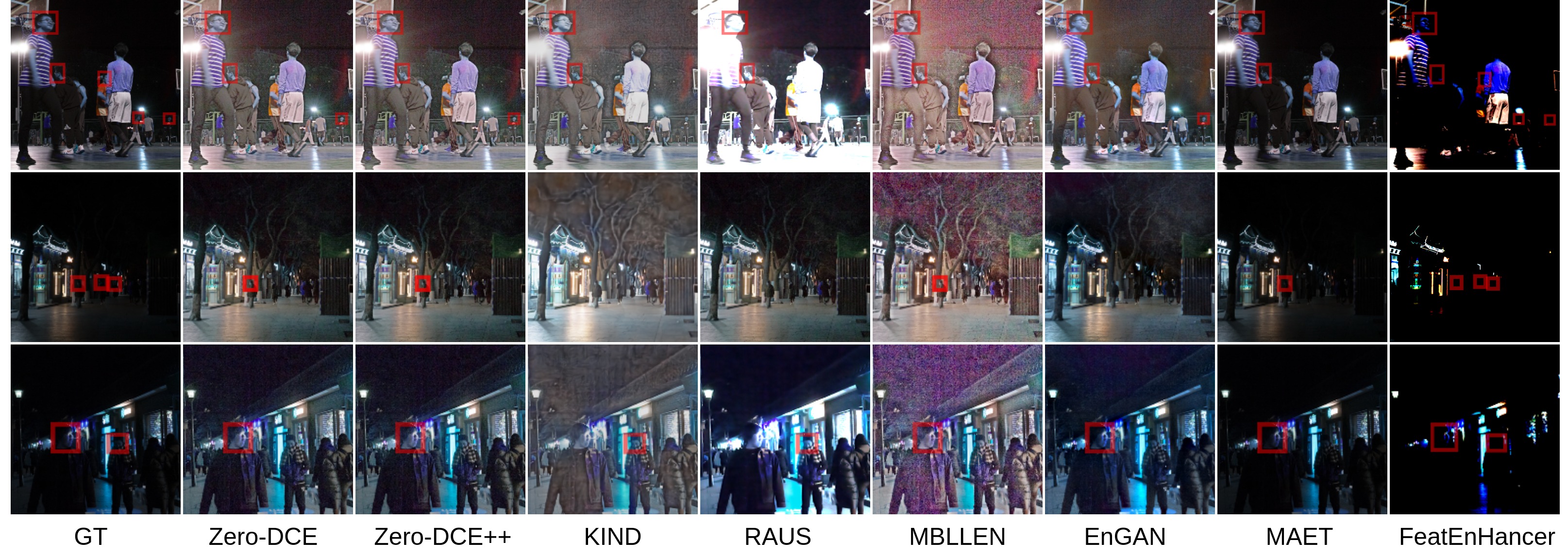}
   \caption{\textbf{Visual comparison of FeatEnHancer with several LLIE approaches and an existing state-of-the-art dark object detection method on the DARK FACE dataset}. Zoom in for the best view.}
\label{fig:qual_darkface_bigger}
\end{center}
\end{figure*}

\begin{table*}
 \begin{adjustbox}{width=\textwidth}
\centering
 \begin{tabular}{c|c c c c c c c c c c c c c c c c c c c c c c c c >{\columncolor{gray!20}}c >{\columncolor{gray!20}}c c c c c } 
    \toprule
    \multirow{2}{*}{Method} &
      \multicolumn{2}{c}{Bicycle} &
      \multicolumn{2}{c}{Boat} &
      \multicolumn{2}{c}{Bottle} &
      \multicolumn{2}{c}{Bus} &
      \multicolumn{2}{c}{Car} &
      \multicolumn{2}{c}{Cat} &
      \multicolumn{2}{c}{Chair} &
      \multicolumn{2}{c}{Cup} &
      \multicolumn{2}{c}{Dog} &
      \multicolumn{2}{c}{Motorbike} &
      \multicolumn{2}{c}{People} &
      \multicolumn{2}{c}{Table} &
      \multicolumn{2}{c}{\textbf{AP50}} &
      \multicolumn{2}{c}{\textbf{AP75}} &
      \multicolumn{2}{c}{\textbf{mAP}} \\
      & { Ret} & {FQ} & { Ret} & {FQ} & { Ret} & {FQ}
      & { Ret} & {FQ} & { Ret} & {FQ} & { Ret} & {FQ}
      & { Ret} & {FQ} & { Ret} & {FQ} & { Ret} & {FQ}
      & { Ret} & {FQ} & { Ret} & {FQ} & { Ret} & {FQ}
      & { Ret} & {FQ} & { Ret} & {FQ} & { Ret} & {FQ} \\
      \midrule
    Baseline & 50.4  & 57.1 & 40.9  & 42.0 & 34.2  & 45.8 & 73.1  & 73.9 & 57.4  & 56.4 & 45.2  & 41.2 & 42.0  & 39.4 & 46.2  & 46.5 & 50.6  & 52.2 & 40.2  & 36.7 & 41.6  & 42.8 & 33.5  & 30.5 & 72.1 & 74.5 & 51.4 & 44.1 & 46.3 & 47.0 \\

   RAUS~\cite{raus} & 49.4    & 53.9 & 37.9  & 43.8 & 33.6  & 42.5 & 68.3    & 69.5 & 53.6  & 52.9 & 41.5  & 42.6 & 40.9    & 47.2 & 41.0  & 47.7 & 48.5  & 48.1 & 37.4    & 39.7 & 39.8  & 43.9 & 33.3 & 44.3 & 64.7 & 77.0 & 44.1 & 49.0 & 44.0 & 48.1 \\ 

    KIND~\cite{kind} & 49.4  & 55.6 & 38.8  & 46.0 & 34.5  & 46.1 & 72.4  & 71.7 & 56.5  & 57.8 & 41.6  & 48.0 & 40.1  & 48.4 & 44.6  & 56.0 & 50.8  & 51.8 & 39.0  & 44.3 & 41.0  & 45.1 & 32.3  & 47.4 & 70.7 & 80.5 & 49.6 & 57.2 & 45.1 & 51.5 \\

    Zero-DCE++~\cite{zero-dce++} & 50.0  & 53.4 & 39.8  & 45.0 & 34.1  & 44.4 & 72.2  & 71.4 & 56.6  & 55.1 & 41.7  & 46.9 & 41.0  & 45.0 & 44.2  & 47.6 & 50.2  & 50.4 & 40.3  & 44.7 & 40.4  & 41.7 & 32.2  & 44.4  & 70.3 & 79.5 & 50.1 & 49.2 & 45.2 & 49.2 \\	

    EnGAN~\cite{enlightengan} & 49.5  & 55.1 & 39.9  & 47.2 & 33.7  & 43.3 & 72.6  & 74.5 & 56.1  & 57.7 & 42.0  & 46.9  & 40.3  & 49.2 & 43.1  & 55.4 & 50.1  & 53.1 & 38.8  & 45.0 & 40.7  & 45.8 & 31.6  & 49.1 & 70.4 & 80.0 & 49.7 & 58.7 & 44.9 & 51.9 \\

    MBLLEN~\cite{mebbln} & 50.1  & 55.4 & 38.0 & 45.0 & 33.8  & 47.2 & 72.6  & 72.6 & 57.3 & 59.6 & 41.7  & 46.5 & 41.4  & 46.6 & 43.5 & 52.6 & 49.8  & 51.9 & 40.6  & 45.7 & 41.0 & 46.1 & 32.5  & 47.7 & 70.6 & 80.0 & 49.0 & 58.3 & 45.1 & 51.0 \\

    Zero-DCE~\cite{zero-dce} & 50.8  & 55.8 & 39.6 & 47.0 & 34.9 & 45.2 & 73.5  & 73.0 & 56.7 & 59.0 & 40.2 & 46.8 & 41.0  & 48.1 & 44.1 & 53.9 & 50.0 & 52.9 & 39.5  & 47.4 & 40.8 & 46.5 & 32.3 & 48.1 & 71.0 & 80.6 & 49.8 & 56.7 & 45.2 & 52.0\\
    
    MAET~\cite{zero-dce} & 50.8  & 56.2 & 39.8 & 47.8 & 35.7 & 45.3 & 74.5  & 73.3 & 56.9 & 59.4 & 40.9 & 46.9 & 41.7  & 48.7 & 44.5 & 54.3 & 50.5 & 53.9 & 39.7  & 47.7 & 40.9 & 46.7 & 32.6 & 48.4 & 71.8 & 81.6 & 49.8 & 56.7 & 45.7 & 52.4\\

     \textbf{FeatEnHancer} &  \textbf{51.0}  & \textbf{60.3} & \textbf{41.6}  & \textbf{49.1} & \textbf{47.6 } & \textbf{53.7} & \textbf{69.8}  & \textbf{78.4} &\textbf{ 54.2 } & \textbf{62.1} & \textbf{47.5 } &\textbf{ 52.1} & \textbf{41.8}  &\textbf{ 51.9} & \textbf{41.0}  &\textbf{ 57.7} & \textbf{34.5}  &\textbf{ 61.2} & \textbf{42.7}  & \textbf{50.8} & \textbf{45.8}  & \textbf{54.4} & \textbf{40.3}  & \textbf{46.9} & \textbf{72.6} & \textbf{86.3} & \textbf{51.4} &\textbf{ 63.6} & \textbf{46.4} & \textbf{56.5}\\
    \bottomrule
  \end{tabular}
   \end{adjustbox}
   \caption{\textbf{Detailed comparison of FeatEnHancer with LLIE approaches and existing state-of-the-art dark object detection method on the ExDark dataset.} Here, Ret and FQ represent RetinaNet and Featurized Query R-CNN, respectively. Results obtained on the commonly used evaluation metrics are highlighted. Our FeatEnHancer consistently boosts the baseline performance and achieves new state-of-the-art results with Featurized Query R-CNN.}
   \label{tab:exdark_detail}
\end{table*}

\begin{figure*}
\begin{center}
\includegraphics[width=.9\linewidth]{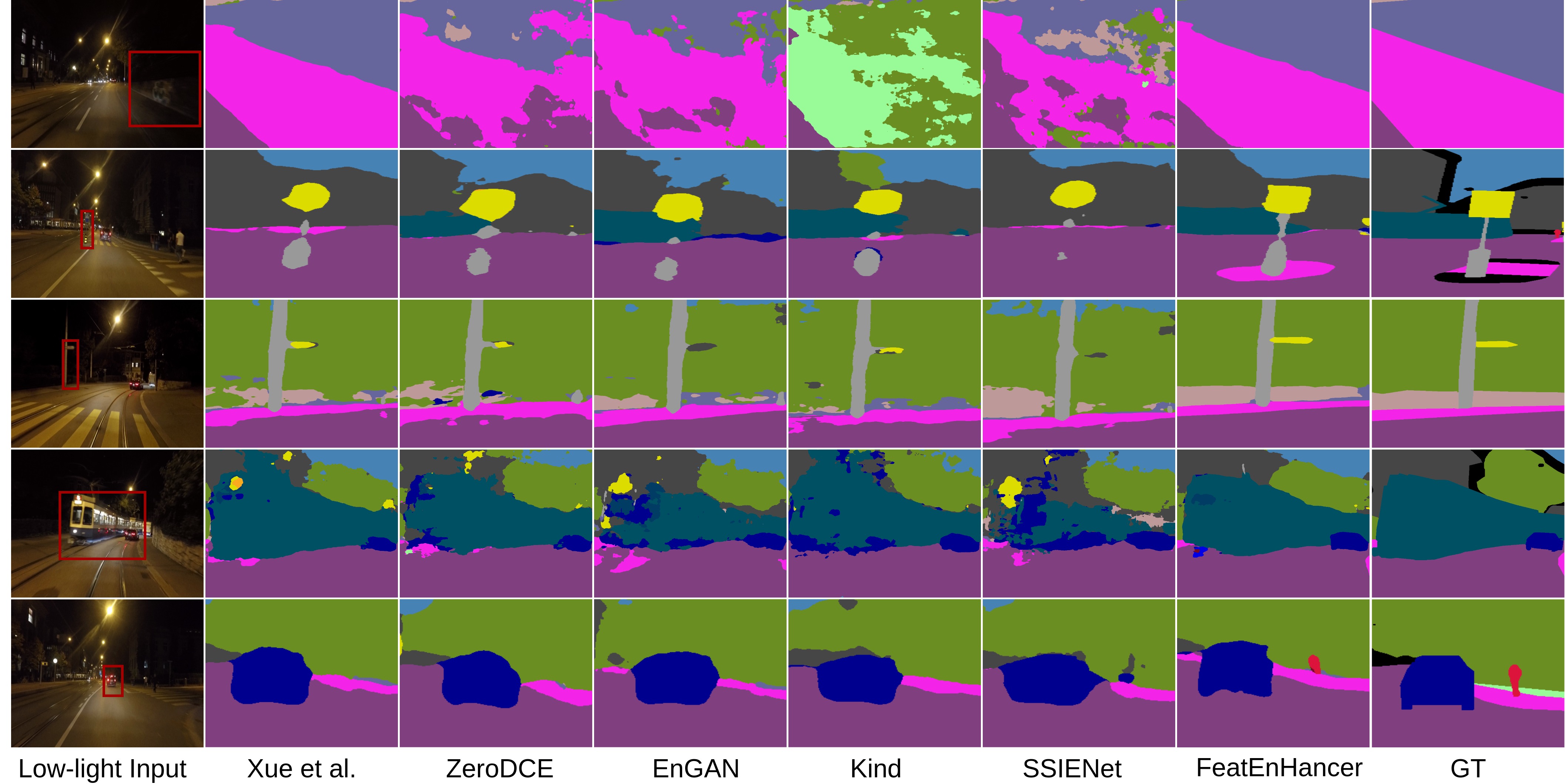}
   \caption{\textbf{Visual comparison of FeatEnHancer with several LLIE approaches and a previous state-of-art nighttime semantic segmentation method on the ACDC dataset}. Zoom in for the best view.}
   \label{fig:qual_ACDC_bigger}
\end{center}
\end{figure*}

\begin{table*}[ht]

\begin{adjustbox}{width=\textwidth}
\footnotesize
\centering
  \begin{tabular}{c|c c c c c c c c c c c c c c c c c | c} 
    Method & RO& SI& BU& WA& FE& PO& TL& TS& VE& TE& SK& PE& RI& CA& TR& TA& BI&\textbf{mIoU} \\
    \hline
    Baseline~\cite{UpperNEt_ECCV_18} & 90.0 & 61.4 & 74.2 & 32.8 & 34.4 & 45.7 & 49.8 & 31.2 & 68.8 & 14.6 & 80.4 & 27.1 & 12.6 & 62.1 & 0.0 & 76.3 & 14.4 & 45.7 \\
    RetinexNet~\cite{retinex_bmvc_18} & 89.4 & 61.0 & 70.6 & 30.1 & 28.1 & 42.4 & 47.6 & 25.7 & 65.8 & 8.6 & 77.3 & 21.5 & 13.8 & 54.8 & 0.0 & 67.4 & 8.2 & 41.9 \\
    DRBN~\cite{DRBN} & 90.5 & 61.5 & 72.8 & 31.9 & 32.5 & 44.5 & 47.3 & 27.2 & 65.7 & 10.2 & 76.5 & 24.2 & 13.2 & 55.4 & 0.0 & 71.1 & 11.9 & 43.3 \\
    FIDE~\cite{FIDE} & 90.0 & 60.7 & 72.8 & 32.4 & 34.1 & 43.3 & 47.9 & 26.1 & 67.0 & 13.7 & 78.0 & 26.5 & 5.8 & 57.1 & 0.0 & 71.0 & 12.4 & 43.4 \\
    KIND~\cite{kind} & 90.0 & 61.0 & 73.2 & 31.9 & 32.8 & 43.5 & 42.7 & 27.7 & 65.5 & 13.3 & 77.4 & 22.8 & 8.1 & 55.1 & 0.0 & 74.5 & 11.5 & 43.0 \\
    EnGAN~\cite{enlightengan} & 89.7 & 58.9 & 73.7 & 32.8 & 31.8 & 44.7 & 49.2 & 26.2 & 67.3 & 14.2 & 77.8 & 25.0 & 10.6 & 59.0 & 0.0 & 71.2 & 7.8 & 43.8 \\
    Zero-DCE~\cite{zero-dce} & 90.6 & 59.9 & 73.9 & 32.6 & 31.7 & 44.3 & 46.2 & 25.8 & 67.2 & 14.6 & 79.1 & 24.7 & 7.7 & 59.4 & 0.0 & 66.8 & 13.9 & 43.4 \\
    SSIENet~\cite{ssie_net_arxiv_20} & 89.6 & 59.3 & 72.5 & 29.9 & 31.7 & 45.4 & 43.9 & 24.5 & 66.7 & 10.6 & 78.3 & 22.8 & 0.2 & 52.6 & 0.0 & 71.1 & 5.4 & 41.4 \\
    Xue~\etal\cite{see_understand_dark_22_ACMM} & 93.2 & 72.6 & 78.4 & 43.8 & 46.5 & 48.1 & 51.1 & 38.8 & 68.6 & 14.9 & 79.1 & 21.9 & 2.2 & 61.6 & \textbf{5.2} & 85.2 & 36.1 & 49.8 \\
     \rowcolor{gray!20} \textbf{FeatEnHancer} &  \textbf{94.0}	& \textbf{75.1} & \textbf{78.6} & \textbf{44.9} & \textbf{41.6} & \textbf{53.9} & \textbf{66.0} & \textbf{49.9} & \textbf{71.2} & \textbf{15.1} & \textbf{82.7} & \textbf{45.3} & \textbf{10.2} & \textbf{72.5} & {0.0} & \textbf{89.5} & \textbf{43.0} & \textbf{54.9} \\
\bottomrule
 \end{tabular}
 \end{adjustbox}
 \caption{\textbf{Comparing FeatEnHancer with LLIE approaches and existing state-of-the-art nighttime segmentation method on the ACDC dataset.} For brevity, we represent classes \{road, sidewalk, building, wall, fence, pole, traffic light, traffic sign, vegetation, terrain, sky, person, rider, car, truck, train, bicycle\} with \{RO, SI, BU, WA, FE, PO, TL, TS, VE, TE, SK, PE, RI, CA, TR, TA, BI\}, respectively. Our FeatEnHancer yields remarkable improvements in the baseline, producing new state-of-the-art results.}
 \label{tab:results_acdc_complete}
\end{table*}

\section{Results and Discussion}
\label{sec:results}

\subsection{Detailed Results on ExDark}
\label{sec:results_exdark}
Table~\ref{tab:exdark_detail} summarizes the exhaustive quantitative analysis, comparing our FeatEnHancer with several LLIE approaches, including RAUS~\cite{raus}, KIND~\cite{kind}, EnGAN~\cite{enlightengan}, MBLLEN~\cite{mebbln}, Zero-DCE~\cite{zero-dce}, Zero-DCE++~\cite{zero-dce++}, and state-of-the-art dark object detection method MAET~\cite{maet} on the ExDark dataset~\cite{exdark-dataset}. It is evident that our FeatEnHancer yields impressive improvements using both object detection frameworks. Furthermore, we exhibit a comprehensive qualitative analysis in Fig.~\ref{fig:qual_exdark_bigger}. Despite the visually unappealing images, the detector equipped with our FeatEnHancer produces accurate detections compared to other LLIE and existing state-of-the-art methods. By looking at the first row of Fig.~\ref{fig:qual_exdark_bigger}, note that while all methods detect the bigger boat, they all miss the smaller boat. However, our FeatEnHancer, enriched with hierarchical multi-scale features, conveniently detects both of the boats. Similarly, bigger and smaller cars are accurately detected by our method in the same figure. These architectural innovations contribute to remarkable improvements in the baseline and deliver new state-of-the-art mAP$_{50}$ of 86.3 on the ExDark dataset with Featurized Query R-CNN.

\subsection{Detailed Results on DARK FACE}
\label{sec:results_darkface}
We demonstrate a qualitative comparison of our FeatEnHancer with existing LLIE methods and existing state-of-the-art dark object detection method on the DARK FACE dataset in Fig.~\ref{fig:qual_darkface_bigger}. Albeit the darkness and tiny faces, our FeatEnHancer provides strong visual cues to the detector and brings maximum gains in the baseline compared to other methods.

\subsection{Detailed Results on ACDC}
\label{sec:results_ACDC}
In Table~\ref{tab:results_acdc_complete}, we present a detailed quantitative analysis, comparing our FeatEnHancer with several LLIE approaches, including RetinexNet~\cite{retinex_bmvc_18}, DRBN~\cite{DRBN}, FIDE~\cite{FIDE}, KIND~\cite{kind}, EnGAN~\cite{enlightengan}, Zero-DCE~\cite{zero-dce}, SSIENet~\cite{ssie_net_arxiv_20}, and prior state-of-the-art nighttime semantic segmentation method Xue~\etal~\cite{see_understand_dark_22_ACMM} on the ACDC dataset~\cite{ACDC_dataset_ICCV_21}. The results show that our FeatEnHancer generates powerful semantic representations, producing significant boosts in the baseline performance and leading to a new state-of-the-art mIoU of 54.9. Moreover, we illustrate a detailed visual comparison in Fig.~\ref{fig:qual_ACDC_bigger}. Note that our method produces accurate segmentations in both cases of small objects, such as traffic signs (second row) and larger objects, such as trains and terrains (first and fourth row). These extra gains demonstrate the effectiveness of the hierarchical multi-scale feature learning and scale-aware attentional feature aggregation schemes in the proposed method.

\subsection{Analysing Features from RetinaNet and Featurized Query R-CNN}
\label{sec:retina_fqrcnn}
\begin{figure*}
\begin{center}
\includegraphics[width=\linewidth]{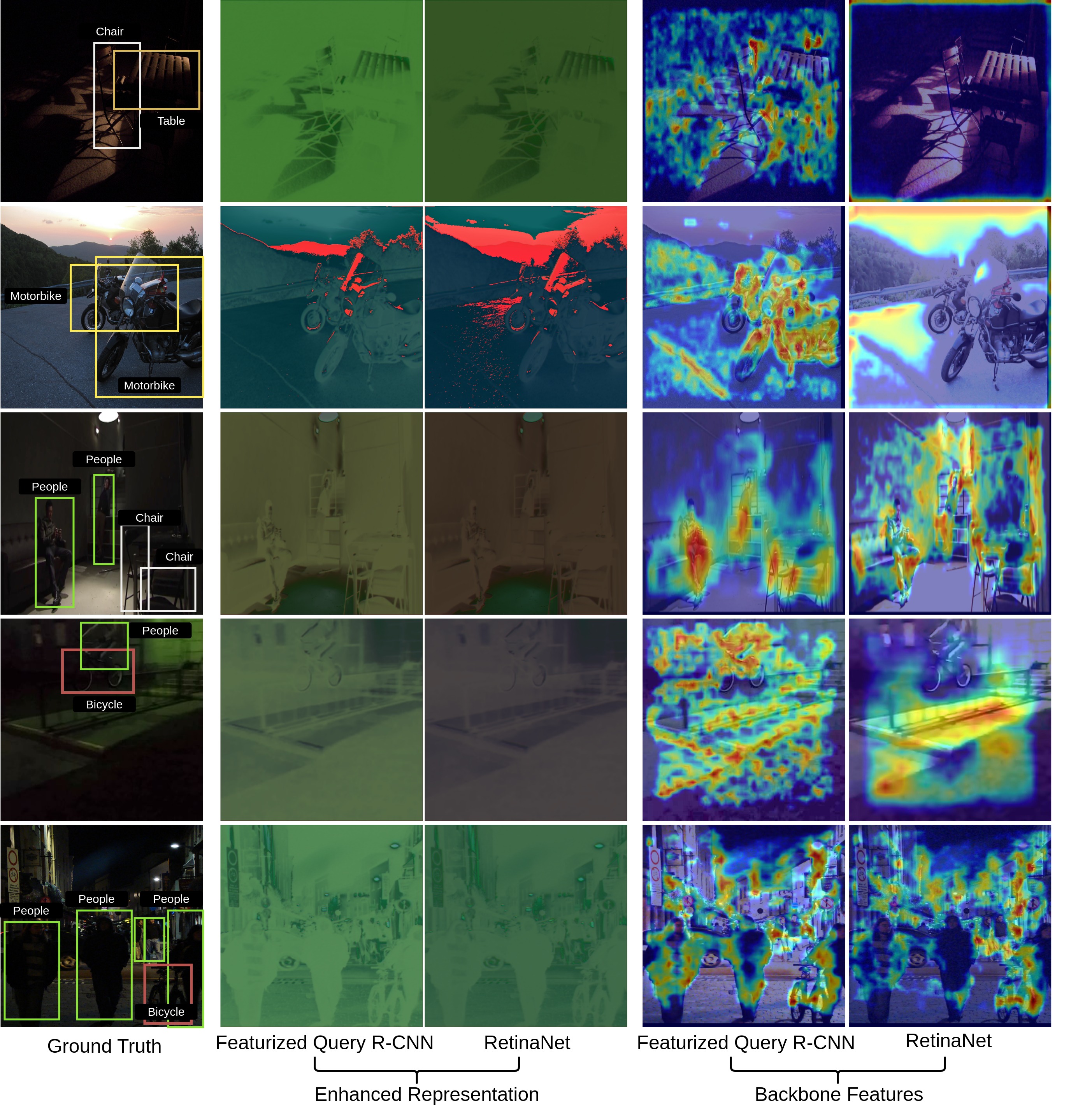}
   \caption{\textbf{Visualizing enhanced hierarchical representation and backbone features from RetinaNet and Featurized Query R-CNN}. We visualize the output of the final enhanced representation achieved after aggregating multi-scale hierarchical features. For backbone features, we employ gradcam~\cite{gradcam} and illustrate learned features from the last \textit{Res4} block in the ResNet-50 backbone network. }
   \label{fig:retina_fq_diff_vis}
\end{center}
\end{figure*}

During experiments with RetinaNet and Featurized Query R-CNN on ExDark and DARK FACE datasets, we observe that our FeatEnHancer brings inferior improvements with RetinaNet (+0.5 mAP$_{50}$), compared to Featurized Query R-CNN ( +11.8 mAP$_{50}$) (see Table~\ref{tab:exdark_detail}). Therefore, we visualize both the learned hierarchical representations of our FeatEnHancer and feature activations from the backbone network (\textit{Res4 block}) in Fig.~\ref{fig:retina_fq_diff_vis}. Note that an identical backbone network of ResNet-50 is employed with both object detectors. From the figure, one can see that with RetinaNet, weaker representations are produced by our method, leading to suboptimal feature extraction, causing inferior gains. On the other hand, with an improved detector like Featurized Query R-CNN, our FeatEnHancer produces more meaningful representations, enabling subsequent backbone networks to extract suitable features. This behaviour is aligned with our network design which is optimized using a task-related loss function. Therefore, our FeatEnHancer can be integrated with advanced downstream vision methods to achieve substantial gains.

\subsection{Comparing FEN in FeatEnHancer with DCENet in Zero-DCE~\cite{zero-dce, zero-dce++}}
\label{sec:fen_against_dce}
\begin{figure*}
\begin{center}
\includegraphics[width=\linewidth]{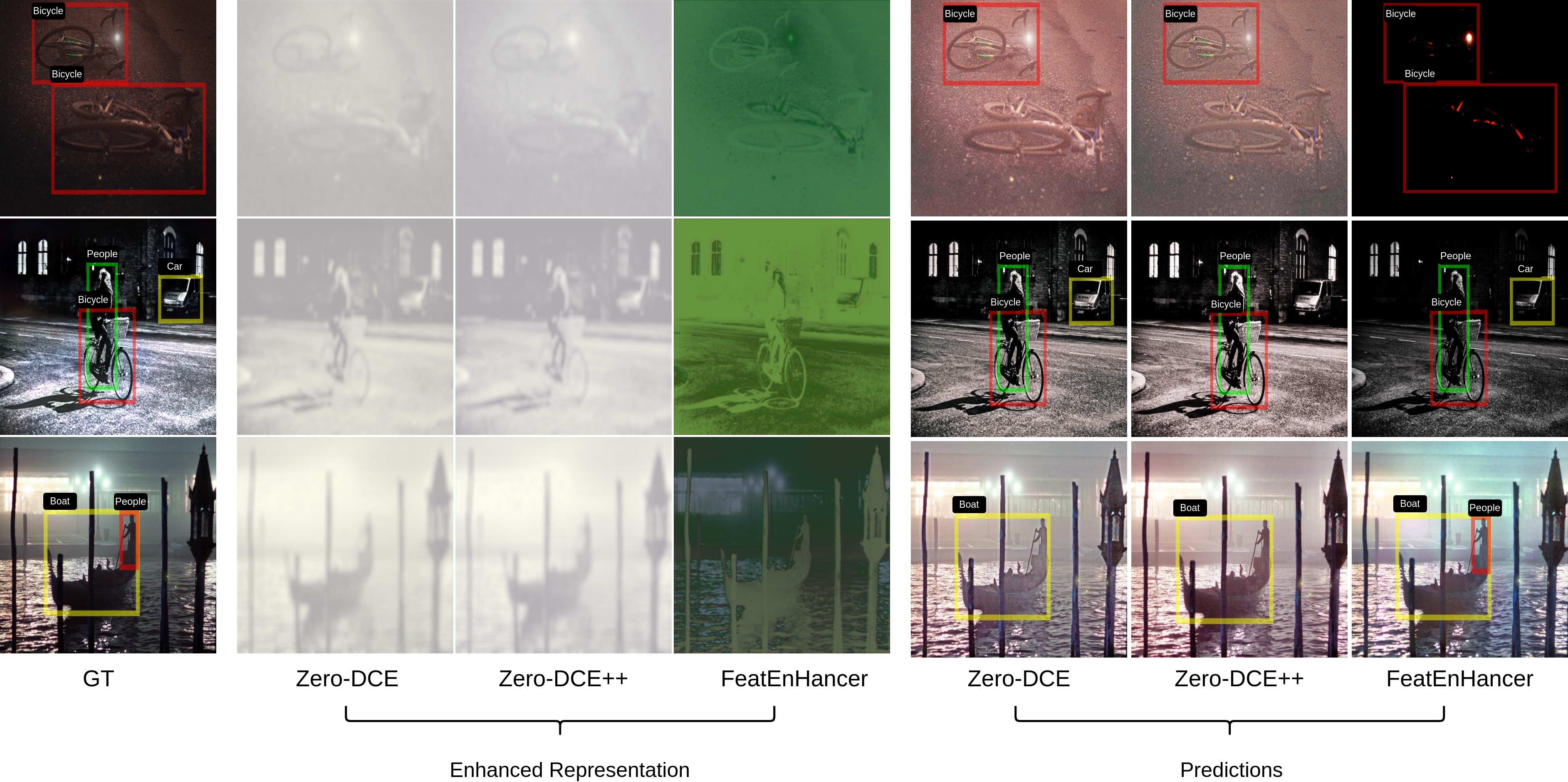}
   \caption{\textbf{Visual comparison between our feature enhancement network and DCENet proposed in~\cite{zero-dce, zero-dce++}}. For comparison, we illustrate the learned mean of all eight curves from DCENet in Zero-DCE and Zero-DCE++. For our FeatEnHancer, we visualize the learned aggregated hierarchical feature representation. All methods are incorporated with Featurized Query R-CNN and trained on the ExDark dataset. }
   \label{fig:dce_vs_ours}
\end{center}
\end{figure*}

Our intra-scale feature enhancement network~(FEN) is inspired by the enhancement network DCENet employed in~\cite{zero-dce} and~\cite{zero-dce++}. However, we incorporate several modifications, such as learning feature enhancement at multiple scales and scale-aware attentional feature aggregation schemes. For direct comparison, we exhibit the learned enhanced representations from both Zero-DCE, Zero-DCE++ and our FeatEnHancer in Fig.~\ref{fig:dce_vs_ours}. All three visualizations are achieved from identical experimental settings on the ExDark validation set. It is obvious that the proposed modification in our FEN produces more meaningful semantic representations compared to DCENet employed in~\cite{zero-dce,zero-dce++}. Both Zero-DCE and Zero-DCE++ produce false negatives, such as missing bicycles and people in the first and third rows, respectively. However, the scale-aware enhancement in our FeatEnHancer leads to accurate detections.

\subsection{Comparison with Domain Adaptation Methods}
\label{sec:domain_adaptation}
Recently some methods~\cite{MS_YOLO_ICIP21, image_adaptive_YOLO_AAAI_22, detection_driven_DENet, domain_adaptive_yolo} have exploited the unsupervised domain adaption scheme to improve object detection in challenging environments. These methods are mainly designed to tackle object detection in harsh weather conditions such as foggy weather. Furthermore, for training, they require pre-training on large synthetic datasets (source dataset) labelled with the same classes that match the classes of the target dataset. Therefore, we refrain from including these works~\cite {MS_YOLO_ICIP21, image_adaptive_YOLO_AAAI_22, detection_driven_DENet, domain_adaptive_yolo} when comparing performance on dark object detection in Table~\ref{tab:exdark_detail}. Nevertheless, for direct comparison with results reported in~\cite{detection_driven_DENet}, we incorporate our FeatEnHancer in the identical experimental settings\footnote{\url{https://github.com/NIvykk/DENet}} (YOLOv3~\cite{yolov3} as a baseline detector) and use the same 10 categories that match with Pascal VOC dataset~\cite{pascalvoc_15}. We present the results of this experiment in Table~\ref{tab:compare_uda}. Note that our FeatEnHancer surpasses the previous best method (DE-YOLO~\cite{detection_driven_DENet} that leverages Laplacian Pyramid~\cite{burt1987laplacian} to generate multi-scale features) in this specific experimental setting and reaches the mAP$_{50}$ of 53.70. 

\begin{table}[ht]
\begin{tabular}{c|c}
   \textbf{Methods} & \textbf{mAP$_{50}$}\\
    \hline
    Baseline~\cite{yolov3} & 43.02\\
    MS-DAYOLO~\cite{MS_YOLO_ICIP21} & 44.25\\
    DAYOLO~\cite{domain_adaptive_yolo} &44.62\\
    DSNet~\cite{DSNet_weather_PAMI21}& 45.31\\
    MAET~\cite{maet} & 47.10\\
    IA-YOLO~\cite{image_adaptive_YOLO_AAAI_22} & 49.53\\
    DE-YOLO~\cite{detection_driven_DENet} & 51.51\\
    \rowcolor{gray!20} FeatEnHancer & \textbf{53.70}\\
    \bottomrule
\end{tabular}
\centering
\caption{\textbf{Quantitative results of our FeatEnHancer and existing UDA methods on the ExDark dataset.} For direct comparison, an identical framework of YOLOv3 as the baseline is adopted. Only 10 classes that match with the Pascal VOC dataset~\cite{pascalvoc_15} are used.}
\label{tab:compare_uda}
\end{table}

\subsection{Performance-Efficiency Tradeoff}
\label{sec:comp_efficiency}
Table~\ref{tab:params} compares the performance-efficiency tradeoff between the proposed FeatEnHancer and LLIE methods and task-specific previous state-of-the-art approaches. While Xue~\cite{see_understand_dark_22_ACMM} and Zero-DCE~\cite{zero-dce} contain fewer parameters, they bring sub-optimal gains to the baseline method. On the contrary, with a slightly more number of parameters, our FeatEnHancer demonstrates generalizability and robustness in 4 different downstream vision tasks under low-light conditions. \textit{Nevertheless, it is worth mentioning that most of the parameters come from the proposed scale-aware attentional feature aggregation module in our FeatEnHancer. We believe that instead of traditional attention mechanism~\cite{attn_all_need_Neurips_17}, incorporating an optimized attentional scheme such as~\cite{Linear_array_network_enhancement} will further reduce the parameters in our FeatEnHancer. We leave this for future works to explore.}

\begin{table}[ht]
\begin{tabular}{c|c|c|c}
   \textbf{Methods} & \textbf{\#Params} & \textbf{mAP$_{50}$} & \textbf{mIoU}\\
    \hline
    MBLLEN~\cite{mebbln} & 450K& 80.0& -\\
    KIND~\cite{kind} & 8M& 80.5& 43.0\\
    EnlightenGAN~\cite{enlightengan} & 9M& 80.0& 43.8\\
    Zero-DCE~\cite{zero-dce}& {79K}& 80.6& 43.4\\
    Xue et. al~\cite{see_understand_dark_22_ACMM} &\textbf{14K}& - & 49.8\\
    MAET~\cite{maet} & 40M& 81.6& -\\
    \rowcolor{gray!20} FeatEnHancer & 138K & \textbf{86.3}& \textbf{54.9}\\
    \bottomrule
\end{tabular}
\centering
\caption{\textbf{Comparing the number of parameters in the FeatEnHancer against LLIE approaches and task-specific state-of-the-art methods.} \#Params is the number of parameters, K and M denote thousands and millions, respectively. mAP$_{50}$ is computed on the ExDark dataset, and mIoU values are taken from the ACDC dataset. We take top competitors of the ExDark and ACDC datasets for comparison. For ExDark, results from Featurized Query R-CNN are adopted.}
\label{tab:params}
\end{table}

\end{document}